\documentclass[conference]{IEEEtran}

\usepackage{times}

\usepackage{graphicx}
\usepackage{booktabs}
\usepackage{multirow}
\usepackage{amsmath}
\usepackage{amssymb}
\usepackage{xcolor}
\usepackage{bbding}

\usepackage[numbers]{natbib}
\usepackage{multicol}
\usepackage[bookmarks=true]{hyperref}
\begin{document}

\title{StreamVLA: Breaking the Reason-Act Cycle via Completion-State Gating}




\IEEEoverridecommandlockouts

\author{
    \IEEEauthorblockN{
        Tongqing Chen\textsuperscript{1},
        Hang Wu\textsuperscript{2}, 
        Jiasen Wang\textsuperscript{3},
        Xiaotao Li\textsuperscript{4},
        Lu Fang\textsuperscript{1, $\dagger$}
\thanks{
    \textsuperscript{$\dagger$}Corresponding author: \texttt{fanglu@tsinghua.edu.cn}.
    Other inquiries can be sent to \texttt{ctq24@mails.tsinghua.edu.cn} and \texttt{h.wu@tum.de}.
}
    }
    
    \vspace{1ex}
    
    \IEEEauthorblockA{
        \textsuperscript{1}Tsinghua University
        \textsuperscript{2}Technical University of Munich, 
        \textsuperscript{3}Shanghai University, 
        \textsuperscript{4}Wuhan University
    }
}


%

\maketitle
\begin{abstract}

Long-horizon robotic manipulation requires bridging the gap between high-level planning (System 2) and low-level control (System 1). 
Current Vision-Language-Action (VLA) models often entangle these processes, performing redundant multimodal reasoning at every timestep, which leads to high latency and goal instability. 
To address this, we present \textbf{StreamVLA}, a dual-system architecture that unifies textual task decomposition, visual goal imagination, and continuous action generation within a single parameter-efficient backbone. 
We introduce a \textbf{``Lock-and-Gated''} mechanism to intelligently modulate computation: only when a sub-task transition is detected, the model triggers \textit{slow thinking} to generate a textual instruction and imagines the \textbf{specific visual completion state}, rather than generic future frames. 
Crucially, this completion state serves as a time-invariant goal anchor, making the policy robust to execution speed variations.
During steady execution, these high-level intents are \textit{locked} to condition a \textbf{Flow Matching} action head, allowing the model to bypass expensive autoregressive decoding for \textbf{72\%} of timesteps. 
This hierarchical abstraction ensures sub-goal focus while significantly reducing inference latency. 
Extensive evaluations demonstrate that StreamVLA achieves state-of-the-art performance, with a \textbf{98.5\%} success rate on the LIBERO benchmark and robust recovery in real-world interference scenarios, achieving a \textbf{48\% reduction in latency} compared to full-reasoning baselines.

\end{abstract}



\section{Introduction}
\label{sec:intro}


Generalist robotic agents face a fundamental dilemma: long-horizon tasks require slow, deliberate reasoning to decompose goals and imagine outcomes, yet physical interaction demands fast, reactive control to handle dynamic disturbances. 
Recent Vision-Language-Action (VLA) models, such as RT-2~\cite{rt2} and OpenVLA~\cite{openvla}, attempt to bridge this gap by scaling multimodal pretraining to robot control. 
While they demonstrate impressive generalization, most existing VLAs employ a homogeneous compute strategy: they perform expensive autoregressive reasoning at every control step---regenerating high-level plans and visual features even when the sub-task state remains unchanged. 
This entanglement of reasoning and control leads to significant computational redundancy and restricts the deployment of reasoning-heavy models in streaming, real-time environments.

\begin{figure}[!t]
\centering
\includegraphics[width=\linewidth]{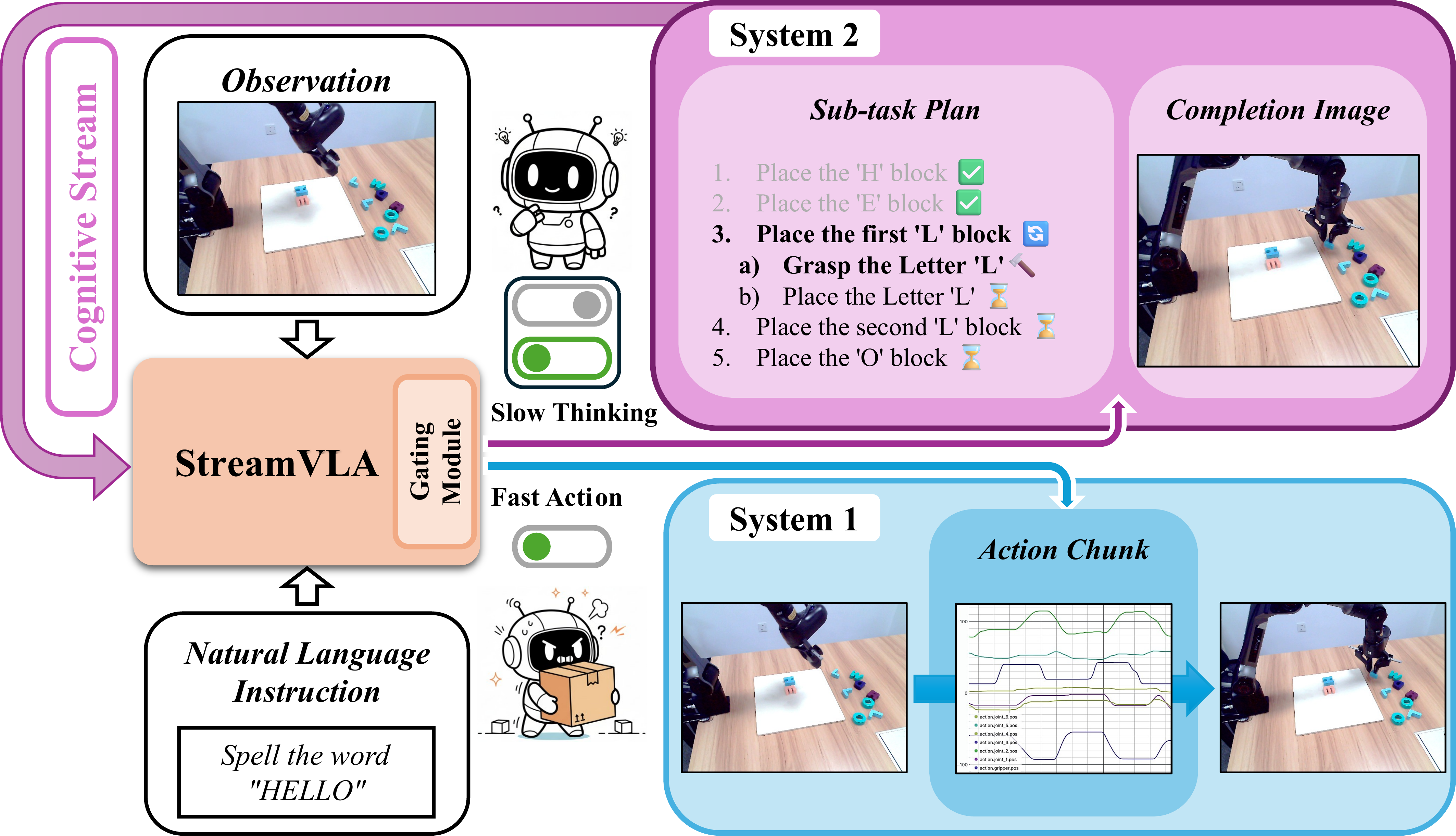}
\caption{
\textbf{StreamVLA Overview.} A dual-system architecture with \textit{Slow Thinking} (purple path) for adaptive sub-task planning and future imagination, and \textit{Fast Action} (blue path) for continuous control. By using predicted future images to gate reasoning, it skips redundant computation when the sub-task is incomplete (72\% of the time), achieving a {48\% reduction in average latency} (244ms $\to$ 128ms).
}
\label{fig:teaser}
\end{figure}

To mitigate latency, recent approaches employ token pruning~\cite{cogvla} or caching strategies~\cite{vla-cache}, often at the cost of losing contextual information.
Others, like CoT-VLA~\cite{cotvla}, incorporate predictive modeling but typically generate future frames at fixed time intervals (e.g., predicting $t+\Delta t$).
We posit that for robotic manipulation, predicting the immediate future is intrinsically flawed due to {temporal misalignment}: execution speed in the real world is stochastic, meaning a fixed-offset prediction may not correspond to any semantically meaningful state.
Instead, effective long-horizon control requires envisioning the {sub-task completion state}---a stable, time-invariant visual anchor that guides the policy until the sub-goal is achieved.
Current architectures lack a mechanism to synchronize this ``goal imagining'' process with the high-frequency control loop.

Inspired by the dual-process theory of human cognition~\cite{thinking}, where \textit{System 2} handles distinct planning and imagination while \textit{System 1} executes effortless motor control, we introduce \textbf{StreamVLA}. 
StreamVLA is a unified framework that disentangles reasoning from execution within a single parameter-efficient backbone. 
Instead of relying on heavy external planner models, we train a lightweight {Infinity-based~\cite{infinity} imagination head} that shares the vast majority of parameters with the action policy.
Our model introduces a ``Lock-and-Gated'' mechanism: only when a sub-task transition is detected, it triggers the heavy \textit{System 2} to generate a textual sub-goal and a specific {visual completion state}.
During the steady execution phase (\textit{System 1}), these high-level intents are \textit{locked} to condition the Flow Matching policy.
This architecture effectively amortizes the cost of reasoning and closes the loop between mental simulation and physical execution.

While hierarchical control and dual-system architectures have been explored in robotics~\cite{cansay, moka}, existing approaches typically rely on {fixed-interval switching}, {textual preconditions}, or {heuristic thresholds} to govern the transition between planning and execution. 
StreamVLA departs from these paradigms by introducing the first {foresight-driven gating mechanism} within a unified VLA. 
Unlike predictive models that generate generic future frames at fixed time steps, our model uses its own {imagined sub-task completion state} as an intrinsic reference. 
The semantic discrepancy between the current observation and this imagined goal dynamically determines whether to persist in ``Fast Action'' mode or trigger ``Slow Thinking'' for re-planning. 
This distinct design ensures that computational resources are allocated based on actual {semantic progress} rather than rigid clock time.

In summary, our contributions are:

\begin{itemize}
    \setlength{\itemsep}{0pt}
    \setlength{\parsep}{0pt}
    \item \textbf{Unified Architecture:} We propose StreamVLA, a dual-system framework that seamlessly integrates Slow Thinking (planning) and Fast Action (control) within a single shared backbone. Unlike prior hierarchical methods that rely on heavy external planners, our design achieves deep reasoning with minimal memory overhead.
    \item \textbf{First Foresight-Driven Gating Mechanism:} We introduce a novel dynamic gating strategy that leverages \textit{imagined sub-task completion states} as the termination signal. To the best of our knowledge, this is the first approach to use self-generated visual foresight to actively modulate the compute allocation of a VLA, skipping redundant reasoning steps with semantic precision.
    \item \textbf{SOTA Performance and Efficiency:} Extensive experiments demonstrate that StreamVLA achieves {State-of-the-Art} results on the LIBERO~\cite{libero} long-horizon benchmark and RoboTwin 2.0~\cite{robotwin2} dynamic tasks. In real-world deployments, our method achieves a {48\% latency reduction} compared to fixed-step baselines, effectively solving the trade-off between reasoning depth and execution speed.
\end{itemize}

\section{Related Work}
\label{sec:related}

\subsection{Vision-Language-Action Foundations}
The convergence of large pretrained vision-language models (VLMs)~\cite{prismatic, paligemma, qwen2.5-vl} with imitation learning has produced a fast-growing class of Vision-Language-Action (VLA) models. 
Early efforts like RT-2~\cite{rt2} fine-tuned VLM backbones directly for discrete action tokenization. 
More recent frameworks, such as OpenVLA~\cite{openvla} and OpenVLA-OFT~\cite{openvlaoft}, adopt autoregressive architectures for generalist policies, while $\pi_{0}$~\cite{pi_zero} and $\pi_{0.5}$~\cite{pi_zero_five} leverage Flow Matching for continuous control. 
Despite their capabilities, these models typically operate in a \textit{dense, episodic manner}: they process each timestep independently without explicit memory of high-level goals. 
While reactive models (like $\pi_{0}$ and $\pi_{0.5}$) are efficient, they lack the {temporal persistence} required for long-horizon coherence. Conversely, reasoning-heavy models perform full-scale multimodal decoding at every step, leading to {computational redundancy} that hinders real-time deployment.

\subsection{Long-Horizon Planning and Visual Foresight}
Long-horizon manipulation necessitates structured reasoning beyond reactive control. 
Chain-of-Thought (CoT) approaches (e.g., CoT-VLA~\cite{cotvla}, ECoT~\cite{ecot}) decompose tasks textually but often lack visual grounding for spatial precision. 
Video prediction models such as UniPi~\cite{unipi} and VLA-OS~\cite{vlaos} guide policies by generating future frames, but typically predict trajectories at {fixed temporal offsets}, which may not align with semantic sub-task boundaries. 
StreamVLA integrates high-fidelity visual generation---leveraging Infinity's bitwise autoregressive paradigm~\cite{infinity, var}---directly into the VLA backbone. 
Crucially, instead of generic video prediction, our model generates {sub-task completion states} as stable goal anchors. This enables a robust, goal-conditioned control loop that is invariant to execution speed.

\subsection{Efficient and Streaming Multimodal Reasoning}
Deploying VLAs on robots requires balancing reasoning depth with real-time latency. 
Existing efficiency strategies largely focus on {token-level optimization}: pruning redundant tokens (CogVLA~\cite{cogvla}), caching KV states (VLA-Cache~\cite{vla-cache}), or compressing model weights (SmolVLA~\cite{smolvla}). 
While effective, these methods often trade off semantic capabilities or require complex memory management (MemoryVLA~\cite{memoryvla}). 
StreamVLA proposes an orthogonal, {system-level strategy}. 
By employing a dynamic gating mechanism, we specifically {bypass the computationally intensive autoregressive heads} (for text and image generation) during steady-state execution, while keeping the shared backbone and action expert active.
This design effectively {amortizes the cost of reasoning}: it reduces the computational burden of logic-heavy models down to the baseline cost of standard reactive VLAs such as $\pi_0$ during execution, granting the system advanced planning capabilities with {negligible overhead} for high-frequency control.

\begin{figure*}[!t]
\centering
\includegraphics[width=0.85\textwidth]{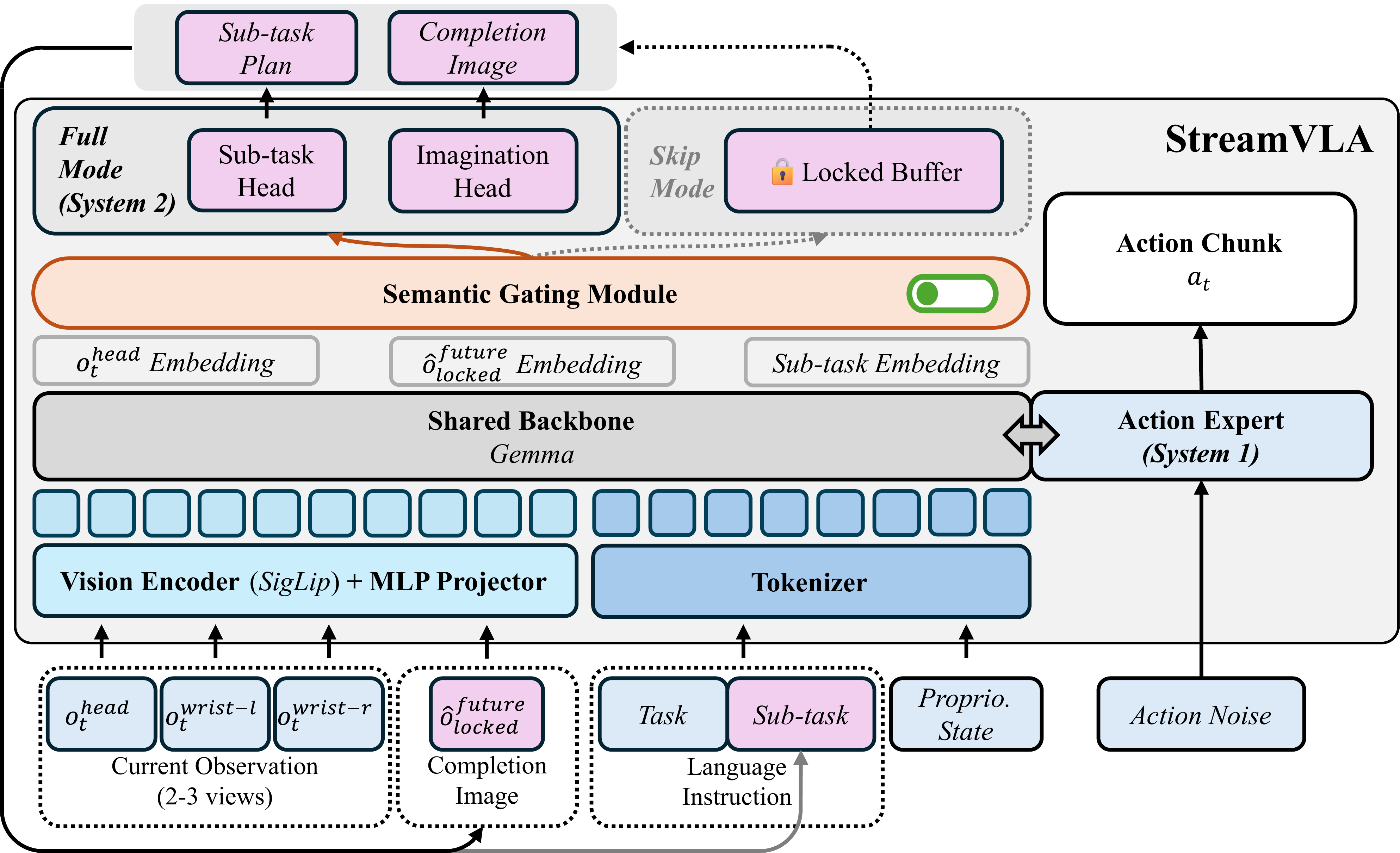}
\caption{\textbf{The StreamVLA Architecture.} 
Our framework unifies sparse reasoning with high-frequency control through a Lock-and-Gated mechanism.
The shared VLA backbone processes multi-view observations. A lightweight Gating Module continuously compares the current state against the locked Completion Image.
If the discrepancy is low (sub-task ongoing), the system operates in Skip Mode, bypassing the computationally expensive heads (System 2) and reusing the cached sub-task plan.
When a transition is detected, Full Mode is triggered to generate a new text plan and visual completion goal.
The Action Expert (System 1) synthesizes precise motor trajectories via Flow Matching, conditioned on the \textit{locked} high-level semantic and visual intents from System 2.
}
\label{fig:architecture}
\end{figure*}

\section{Method}
\label{sec:method}
\subsection{Overview}
\label{sec:overview}

We formulate long-horizon robotic manipulation as a hierarchical decision-making process that alternates between \textit{high-level reasoning} and \textit{low-level control}. 
StreamVLA is designed to operate continuously in a \textbf{streaming setting}, where the agent must efficiently process high-frequency visual observations without incurring the latency of full re-planning at every timestep.

\noindent\textbf{Problem Formulation.} 
Given a natural language instruction $\mathcal{I}$ and a stream of multi-view visual observations $O_t$ (including wrist and head views) and proprioceptive states $\mathbf{p}_t$, the model acts as a policy $\pi$ that outputs a unified tuple at each timestep $t$:
\begin{equation}
    \pi(O_t, \mathbf{p}_t, \mathcal{I}) \rightarrow (\mathcal{S}_t, \mathcal{V}_t, \mathbf{a}_t)
\end{equation}
where:
\begin{itemize}
    \item $\mathcal{S}_t$: A textual sub-task description.
    \item $\mathcal{V}_t$: A \textit{visual completion state} (imagined image) representing the successful outcome of $\mathcal{S}_t$.
    \item $\mathbf{a}_t$: An action chunk $\mathbf{a}_t \in \mathbb{R}^{K \times D}$ for continuous motor control.
\end{itemize}

\noindent\textbf{Unified Architecture.}
StreamVLA unifies these modalities within a single parameter-efficient framework. 
Building upon the $\pi_{0.5}$~\cite{pi_zero_five} foundation, which provides a Vision Encoder $\mathcal{E}_v$ and a unified Transformer Backbone $\mathcal{T}$, we extend the architecture with a {hybrid head design}:
(1) An \textbf{Autoregressive Reasoning Head} (System 2) responsible for generating the high-level plans $\mathcal{S}_t$ and imagining the completion goal $\mathcal{V}_t$; and
(2) A \textbf{Flow Matching Action Head} (System 1) dedicated to generating precise motor trajectories $\mathbf{a}_t$.
Crucially, during execution, the Action Head (System 1) is \textbf{conditioned on} the locked latent representations from System 2, translating high-level intent into low-level control.

\noindent\textbf{Dual-System Inference via Gating.}
Inspired by the efficiency of human cognition~\cite{thinking}, StreamVLA employs a \textbf{Lock-and-Gated mechanism} to modulate computation. 
We distinguish between \textit{Slow Thinking} (generating $\mathcal{S}_t, \mathcal{V}_t$) and \textit{Fast Action} (generating $\mathbf{a}_t$). 
A lightweight Gating Module $\mathcal{G}$ continuously monitors the alignment between the current observation $O_t$ and the previously imagined goal $\mathcal{V}_{prev}$.
\begin{itemize}
    \item \textbf{System 1 Locked:} During steady execution, the autoregressive head is bypassed. The action head is conditioned on the \textit{locked} high-level intents ($\mathcal{S}_{locked}, \mathcal{V}_{locked}$), enabling high-frequency streaming inference.
    \item \textbf{System 2 Triggered:} If a sub-task transition or failure is detected, the full autoregressive head is activated to re-plan ($\mathcal{S}_{new}, \mathcal{V}_{new}$).

\end{itemize}

The following subsections detail the implementation of the shared backbone, the hybrid heads, and the gating dynamics.


\subsection{Future-Completion Visual Imagination}
\label{sec:future-imagination}

Unlike standard video prediction models that generate frames at fixed time intervals~\cite{unipi, vlaos}, our Imagination Head is trained to generate the {visual state of success} for the current sub-task. 
This distinction is critical: execution speed in the real world is stochastic, but the completion state (e.g., ``drawer fully closed'') remains invariant.
By predicting the specific outcome rather than an arbitrary future frame, we provide a robust, visually grounded goal that anchors the policy regardless of temporal variations.

To implement this efficiently, we adapt the {Infinity} architecture~\cite{infinity}---a bitwise autoregressive model with infinite-vocabulary tokenization---as a lightweight decoder head attached to our shared VLA backbone.
Formally, the imagination head $\mathcal{H}_{\text{img}}$ autoregressively generates a predicted completion image $\hat{o}_t^{\text{future}}$ conditioned on the current multimodal context:
\begin{equation}
    \hat{o}_t^{\text{future}} \sim P_{\theta}(\cdot | o_t, \mathbf{p}_t, \mathcal{I}, s_t)
\end{equation}
where $o_t$ is the current observation, $\mathbf{p}_t$ is the proprioceptive state, $\mathcal{I}$ is the global instruction, and $s_t$ is the generated sub-task plan.
We leverage KV-Cache to accelerate the autoregressive decoding, ensuring that the System 2 reasoning step remains computationally tractable.

\noindent\textbf{Fixed-Viewpoint Stability.} 
To ensure consistent progress tracking, we specifically designate the fixed head-mounted camera view for future prediction, rather than predicting all camera views. 
This design choice reduces generation costs and provides a stable, occlusion-free reference frame for the gating mechanism to reliably compute the visual discrepancy between the current state and the goal state.

This generated completion state serves two critical roles in our dual-system framework:
\begin{enumerate}
    \item \textbf{Visual Goal Guidance:} It acts as a \textit{visual prompt} for the Action Head, providing spatial cues about the target object configuration that text alone cannot convey.
    \item \textbf{Gating Reference:} It serves as the ``ground truth'' for the Gating Module. By comparing the live observation $o_t$ against the locked goal $\hat{o}^{\text{future}}_{locked}$, the system can implicitly measure sub-task progress (as detailed in Sec.~\ref{sec:gating}).
\end{enumerate}

\begin{figure}[t]
\centering
\includegraphics[width=0.9\linewidth]{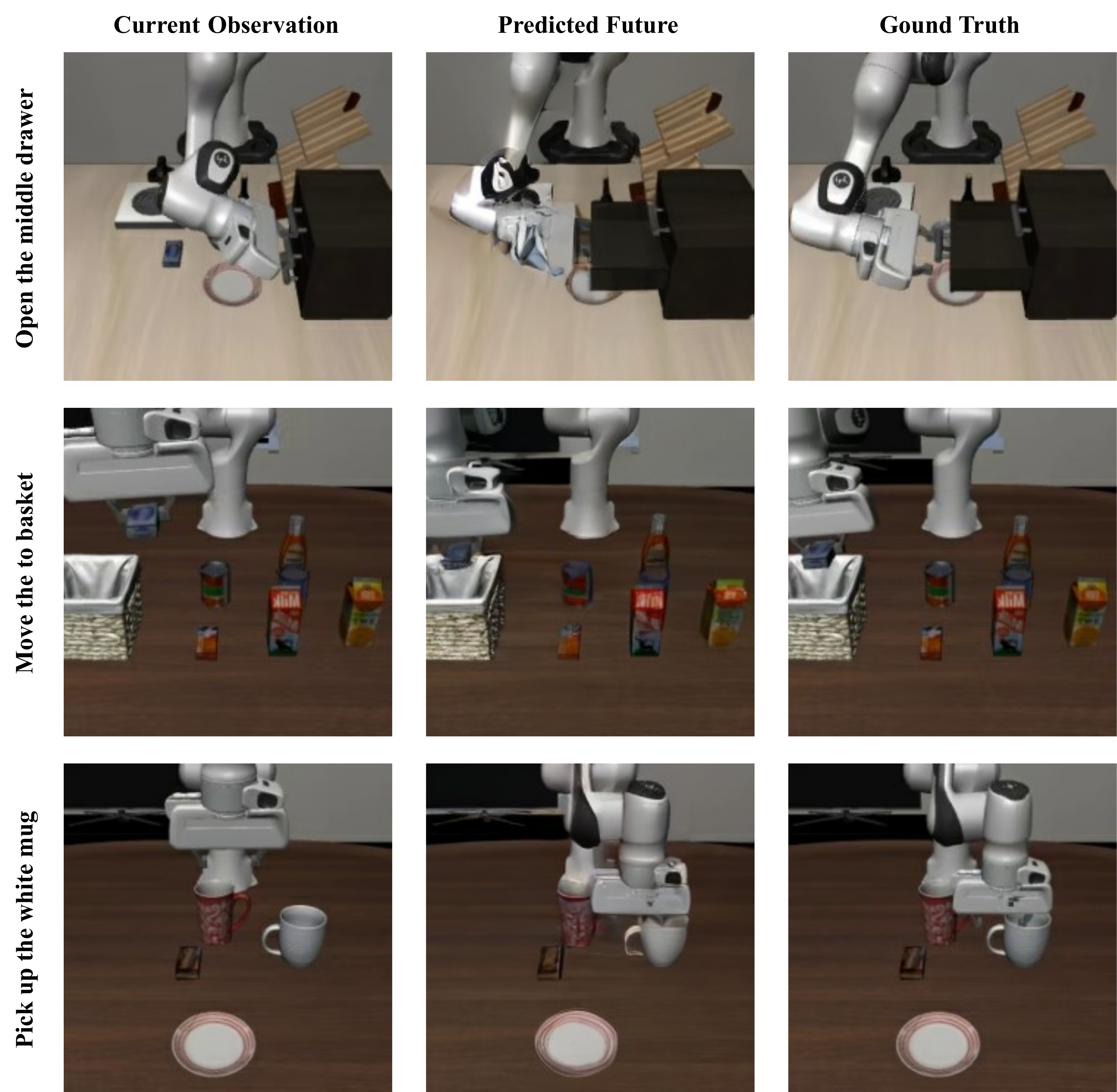}
\caption{\textbf{Future-Completion Visual Imagination Examples.} Three LIBERO tasks showing (left to right): current observation, predicted future completion frame, and ground truth completion state. Unlike methods that predict the next frame ($t+\Delta t$), StreamVLA generates the sub-task completion state (Goal). This goal remains stable throughout the execution of the sub-task, serving as a robust anchor for the gating mechanism.}
\label{fig:future_imagination}
\end{figure}

Figure~\ref{fig:future_imagination} visualizes these generated completion states across various long-horizon tasks.


\subsection{Lock-and-Gated Mechanism}
\label{sec:gating}

To ensure inference efficiency, the Gating Module $\mathcal{G}$ serves as a lightweight {temporal conductor} between System 1 and System 2. 
It adds negligible parameters ($\sim$2\%) yet critically governs the computational flow.

\noindent\textbf{Semantic Discrepancy Estimation.}
At each timestep $t$, the module computes a {Discrepancy Score} $d_t$ representing the semantic distance between the current state and the locked completion goal.
It aligns the current head-mounted observation $o_t^{\text{head}}$ (Query) with the locked future completion state $\hat{o}_{\text{locked}}^{\text{future}}$ (Key/Value) via Cross-Attention, conditioned on the sub-task instruction embedding $e_t = \mathcal{E}_l(s_t)$:
\begin{equation}
    h_t = \text{CrossAttn}(Q=o_t^{\text{head}}, K=\hat{o}_{\text{locked}}^{\text{future}}, V=\hat{o}_{\text{locked}}^{\text{future}}) \oplus e_t
\end{equation}
The fused features $h_t$ are then passed through a lightweight MLP classifier to predict a {Discrepancy Score} $d_t \in [0, 1]$:
\begin{equation}
    d_t = \sigma(\text{MLP}(\text{GlobalPool}(h_t)))
\end{equation}
where $\sigma$ is the sigmoid function. 
A high $d_t$ indicates a significant gap between reality and the goal (i.e., sub-task is ongoing), while a low $d_t$ implies convergence to the completion state.

\noindent\textbf{Gating Logic: To Plan or To Act?}
We employ a threshold-based policy to interpret $d_t$:
\begin{itemize}
    \item \textbf{Phase 1: Execution (High Discrepancy, $d_t > \tau$).} A large $d_t$ implies the robot is visually far from the completion state (i.e., the sub-task is ongoing). The system enters \textbf{Skip Mode}: it bypasses the heavy autoregressive heads, reusing the locked plan ($s_{\text{locked}}, \hat{o}_{\text{locked}}^{\text{future}}$) to drive the Action Expert.
    \item \textbf{Phase 2: Transition (Low Discrepancy, $d_t \leq \tau$).} A small $d_t$ indicates the robot has reached the visual goal (sub-task complete). This triggers \textbf{Full Reasoning Mode}: System 2 activates to generate the new sub-task $s_{t}$ and imagine a new goal $\hat{o}_{t}^{\text{future}}$.
\end{itemize}
We empirically set $\tau = 0.5$. 
At $t=0$, we force a System 2 update to initialize the loop.

\noindent\textbf{Supervision.}
The gating module is trained via binary cross-entropy loss $\mathcal{L}_{\text{gate}}$. 
Ground truth labels are derived from the temporal sub-task boundaries in the expert demonstrations: frames within a sub-task are labeled as $1$ (ongoing/gap), and frames at transition boundaries are labeled as $0$ (completed/no-gap).


\begin{figure}[t]
\centering
\includegraphics[width=0.8\linewidth]{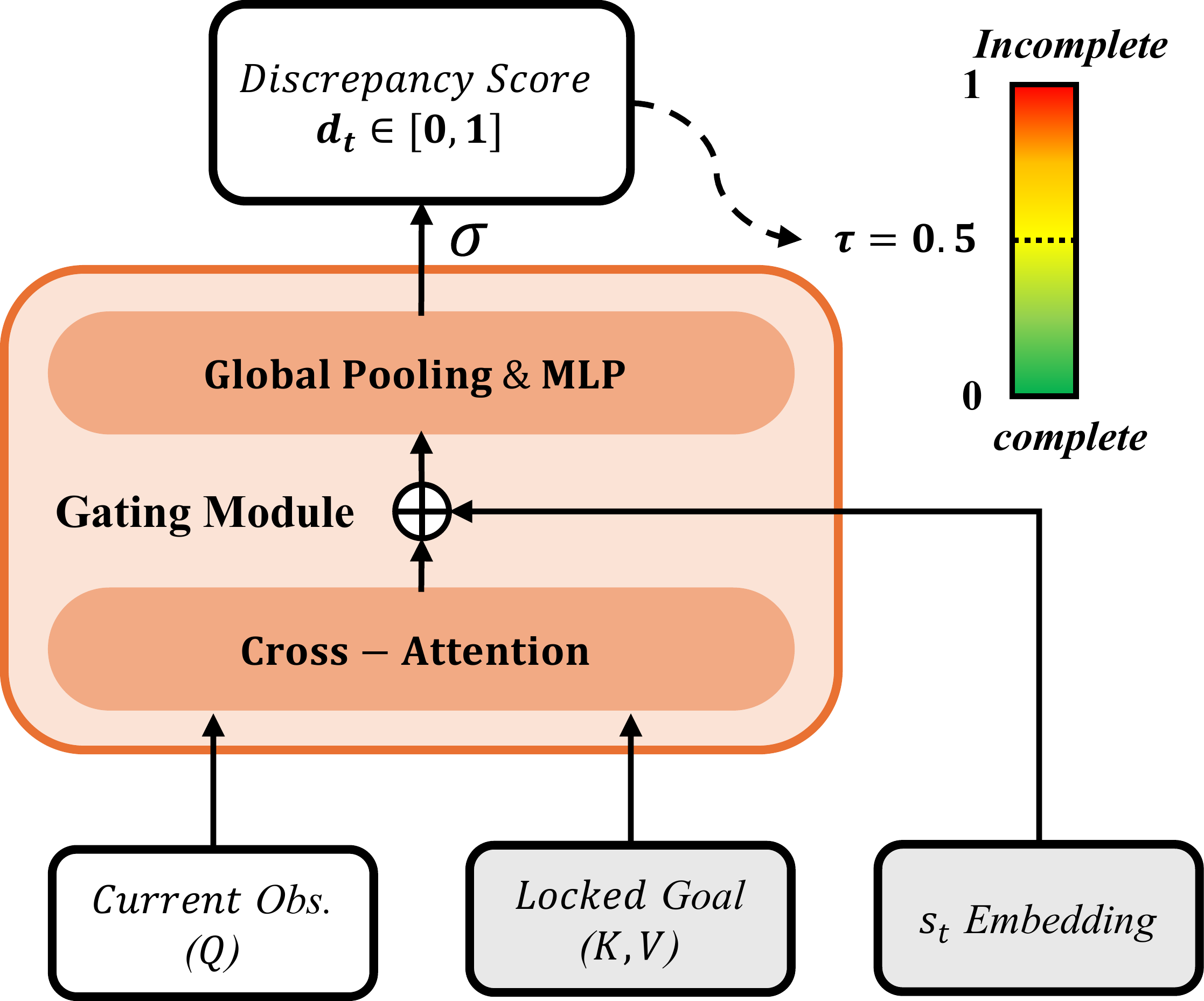}\\[0.3em]
\textbf{(a)} Semantic Gating Module\\[0.5em]
\includegraphics[width=0.8\linewidth]{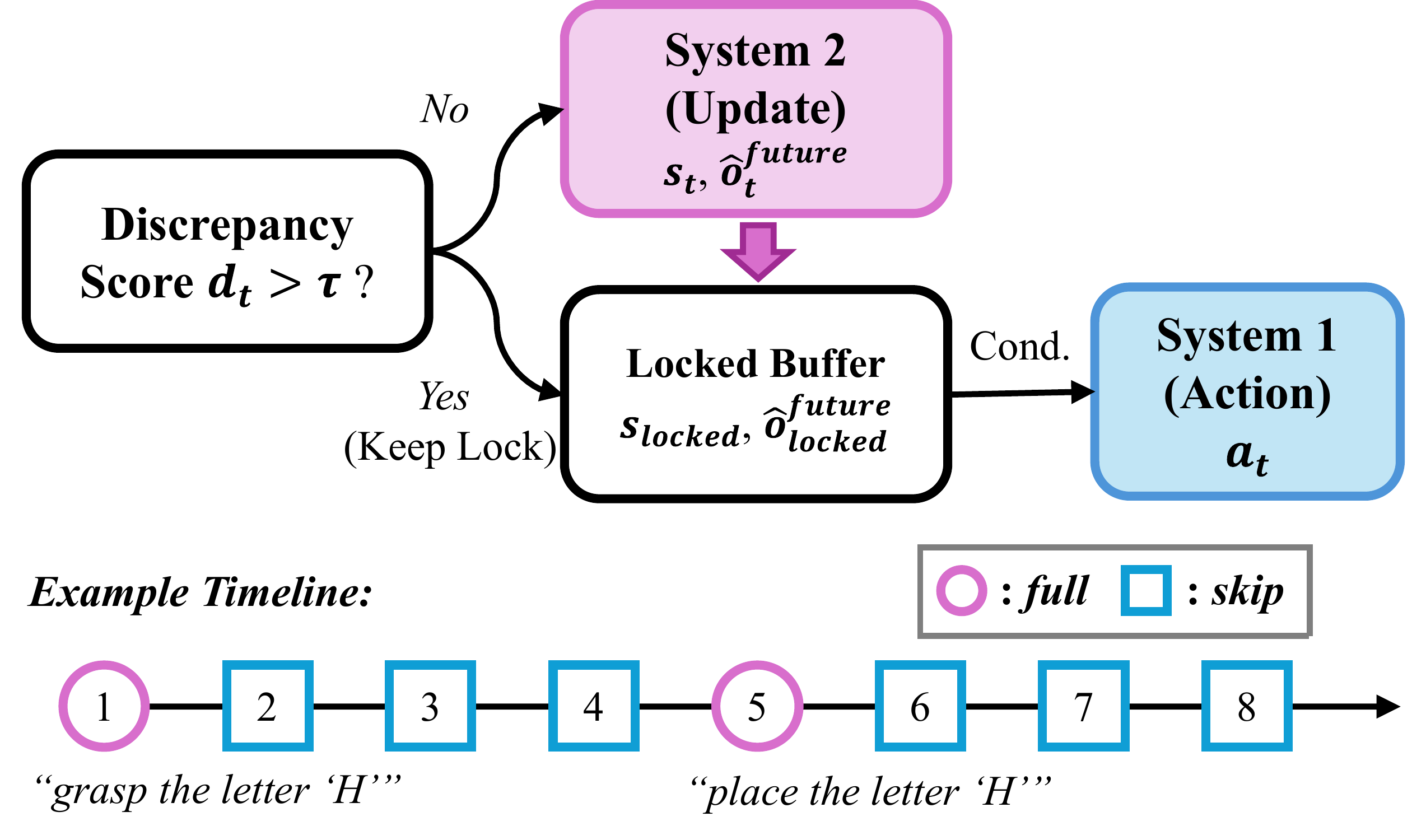}\\[0.3em]
\textbf{(b)} Decision Flow
\caption{\textbf{Lock-and-Gated Mechanism.} 
\textbf{(a)} The Gating Module computes a Discrepancy Score $d_t$ by comparing current observation against the locked goal.
\textbf{(b)} Control Logic: If $d_t > \tau$ (Gap Large), the system \textit{locks} reasoning and executes actions (Skip Mode). If $d_t \leq \tau$ (Gap Closed), it triggers re-planning (Full Mode).
}
\label{fig:gating}
\end{figure}

\subsection{Unified Multi-Modal Output}
\label{sec:unified-output}

StreamVLA integrates these capabilities through a unified transformer architecture. 
The interaction between heads follows a strictly hierarchical structure:

\noindent\textbf{System 1: Flow Matching Action Expert (Dense).}
For motor control, we employ Conditional Flow Matching (CFM)~\cite{flowmatching} rather than diffusion, enabling faster, deterministic trajectory generation.
The Action Expert $\mathcal{H}_{\text{act}}$ predicts a chunk of $K$ actions $\mathbf{a}_t$.

\noindent\textbf{System 2: Autoregressive Planning (Sparse).}
When triggered, two heads generate the high-level intent:
\begin{itemize}
    \item \textbf{Sub-task Head ($\mathcal{H}_{\text{sub}}$):} Generates textual instructions $s_t$ via causal language modeling.
    \item \textbf{Imagination Head ($\mathcal{H}_{\text{img}}$):} Generates the visual completion state $\hat{o}_t^{\text{future}}$ via bitwise autoregressive modeling.
\end{itemize}
These outputs are \textbf{cached} (locked) to guide the low-level controller.

\noindent\textbf{Cross-System Conditioning.}
The core innovation lies in how $\mathcal{H}_{\text{act}}$ utilizes the high-level intent. 
Instead of simple concatenation, we employ a {hierarchical conditioning mechanism}. 
The action head is conditioned on a composite context vector $C_t$:
\begin{equation}
    C_t = [ \mathcal{E}_v(o_t), \mathcal{E}_p(\mathbf{p}_t), \underbrace{\mathcal{E}_l(s_{\text{locked}})}_{\text{Text Goal}}, \underbrace{\mathcal{E}_v(\hat{o}_{\text{locked}}^{\text{future}})}_{\text{Visual Goal}} ]
\end{equation}
This multi-modal conditioning ensures that the synthesized motor commands are informed by:
(1) \textbf{Current Perception} ($o_t, \mathbf{p}_t$) for collision avoidance and alignment;
(2) \textbf{Semantic Intent} ($s_{\text{locked}}$) for task logic;
(3) \textbf{Spatial Anticipation} ($\hat{o}_{\text{locked}}^{\text{future}}$) for precise geometric targeting.
This design allows System 1 to execute complex maneuvers ``blindly'' regarding the high-level plan, relying on the locked goals until the Gating module signals a need for update.


\subsection{Data Annotation}
\label{sec:annotation}

We curate a comprehensive multimodal dataset from three sources: the LIBERO benchmark~\cite{libero}, the RoboTwin 2.0 simulation suite~\cite{robotwin2}, and a self-collected real-world manipulation dataset.

\noindent\textbf{LIBERO and Real-World Data (Hybrid Pipeline).}
Since these datasets lack fine-grained sub-task labels, we adopt a semi-automated annotation pipeline extended from VLA-OS~\cite{vlaos}. 
(1) \textbf{VLM-based Parsing:} We utilize Qwen3-VL-Plus~\cite{qwen3-vl} to generate temporal scene descriptions and Qwen3-Max~\cite{qwen3} to reason about sub-task decomposition based on gripper key-frames.
(2) \textbf{Future Frame Extraction:} Future-completion frames are automatically extracted at the boundary timestamps of each segmented sub-task.
(3) \textbf{Human Verification:} To ensure high-quality supervision for the Gating module, we developed a web-based tool where expert annotators verify and refine the VLM-generated boundaries. This results in a rigorous Chain-of-Thought (CoT) dataset aligned with physical states.

\noindent\textbf{RoboTwin 2.0 (Programmatic Annotation).}
Leveraging the ground-truth simulator states available in RoboTwin, we implement a {programmatic annotator}. 
We define logic rules based on object-goal spatial relations to automatically segment trajectories into sub-tasks and extract precise completion frames. 
This provides noise-free supervision for training the imagination and gating heads without manual intervention.

\subsection{Training Strategy}
\label{sec:training}

StreamVLA is optimized using a multi-task objective that balances continuous control with discrete reasoning.

\noindent\textbf{1. Action Generation ($\mathcal{L}_{\text{act}}$):} 
We employ the Conditional Flow Matching (CFM) objective~\cite{flowmatching}. Given a target action chunk $\mathbf{a}_1$, sampled noise $\mathbf{a}_0 \sim \mathcal{N}(0, I)$, and time $t \in [0, 1]$, we minimize the mean squared error between the model output $v_t$ and the flow vector field:
\begin{equation}
    \mathcal{L}_{\text{act}} = \mathbb{E}_{t, \mathbf{a}_0, \mathbf{a}_1} \left[ || v_t(\phi_t(\mathbf{a}_0), \text{cond}) - (\mathbf{a}_1 - \mathbf{a}_0) ||^2 \right]
\end{equation}

\noindent\textbf{2. Sub-task \& Imagination ($\mathcal{L}_{\text{sub}}, \mathcal{L}_{\text{img}}$):} 
For the autoregressive heads, we use standard cross-entropy loss. $\mathcal{L}_{\text{sub}}$ supervises the text tokens, while $\mathcal{L}_{\text{img}}$ supervises the quantized visual tokens (following Infinity's bitwise modeling):
\begin{equation}
    \mathcal{L}_{\text{gen}} = \lambda_{\text{sub}}\mathcal{L}_{\text{sub}} + \lambda_{\text{img}}\mathcal{L}_{\text{img}}
\end{equation}

\noindent\textbf{3. Gating Supervision ($\mathcal{L}_{\text{gate}}$):} 
We train the gating module using Binary Cross-Entropy (BCE) loss against the ground-truth transition labels derived from our annotation pipeline.

\noindent\textbf{Joint Optimization.}
The total loss is a weighted sum:
\begin{equation}
\mathcal{L}_{\text{total}} = \mathcal{L}_{\text{act}} + \mathcal{L}_{\text{gen}} + \lambda_{\text{gate}}\mathcal{L}_{\text{gate}}
\end{equation}
We empirically set $\lambda_{\text{sub}}=0.1$, $\lambda_{\text{img}}=0.1$, and $\lambda_{\text{gate}}=1.0$ to balance the gradients between the heavy visual generation task and the precise action regression task.

\noindent\textbf{Two-Stage Curriculum.}
To prevent gradient conflict between the newly initialized heads and the pre-trained backbone, we adopt a two-stage strategy:
\begin{itemize}
    \item \textbf{Stage I (Imagination Alignment):} We freeze the VLM backbone and the Action Head, training only the Imagination and Sub-task Heads. This forces the AR heads to learn to extract valid goal representations from the backbone's frozen latent space.
    \item \textbf{Stage II (Full Fine-tuning):} We unfreeze all modules (backbone, action head, AR heads) and perform end-to-end joint training. This allows the backbone to adapt its representations to serve both System 1 (Control) and System 2 (Reasoning) optimally.
\end{itemize}
\section{Experiments}
\label{sec:experiments}

\subsection{Experimental Setup}
\label{subsec:setup}






\noindent\textbf{Hardware Embodiment.} 
We evaluate StreamVLA on a 6-DoF {AgileX Piper} robotic arm across two representative task categories:
\begin{itemize}
    \item \textbf{Fine-Grained Precision Tasks:} Such as the \textit{Letter Insertion} task, which requires sub-centimeter accuracy and complex geometric alignment.
    \item \textbf{Long-Horizon Reasoning Tasks:} Such as the \textit{Spelling Bee} and \textit{Interference Recovery} tasks, which test the model's temporal logic and closed-loop adaptability under dynamic human interference.
\end{itemize}
The system is controlled at a high frequency of {50Hz} via CAN bus to ensure reactive and smooth motion. 
The perception suite consists of a third-person USB Webcam providing a global workspace view and a wrist-mounted RGB camera for close-up visual feedback during manipulation.

\noindent\textbf{Simulation Benchmarks.}
To evaluate StreamVLA across distinct control regimes, we utilize two standard benchmarks:
\begin{itemize}
    \item \textbf{LIBERO~\cite{libero} (Long-Horizon Reasoning):} Comprises four task suites (Spatial, Object, Goal, Long) evaluating generalization across layouts and sequences. 
    \textit{Config:} EEF states + relative pose actions at {10Hz}, chunk $K=10$.
    \item \textbf{RoboTwin 2.0~\cite{robotwin2} (Dynamic Control):} Provides 50 dual-arm tasks with strong domain randomization. 
   \textit{Config:} joint states + relative joint actions at {50Hz}, chunk $K=50$.
\end{itemize}


\noindent\textbf{Training Infrastructure.} 
We train all models with DDP + BF16 on 24 NVIDIA A800 GPUs, using the same combined dataset for fair comparison.

\subsection{Main Results}
\label{subsec:main}

\subsubsection{LIBERO Benchmark Comparison}

As shown in Table~\ref{tab:libero_comparison}, StreamVLA achieves a new state-of-the-art with a {98.5\% average success rate}, surpassing the previous best method (OpenVLA-OFT~\cite{openvlaoft}) by 1.4\% while utilizing significantly fewer parameters (3B vs. 7B).


\noindent\textbf{Robustness in Long-Horizon Tasks.} StreamVLA excels in the \texttt{LIBERO-Long} suite, which requires sequential sub-goal execution. While standard episodic models like OpenVLA and CoT-VLA suffer performance drops of {31\%} and {18.5\%} as the horizon extends, StreamVLA maintains a {96.6\%} success rate (only a {2.6\%} decrease). This demonstrates that our \textit{foresight-driven gating} effectively mitigates goal drift and ensures sub-task coherence, addressing the primary bottleneck of reactive policies.

\noindent\textbf{Parameter Efficiency.}
StreamVLA (3B) outperforms larger baselines (up to 8.5B), suggesting structure can substitute for scale.

\begin{table*}[t]
	\centering
	\footnotesize
	\caption{\textbf{Comparison on LIBERO benchmark.} Success rates (\%) across four task suites. Params in billions. Best in \textbf{bold}, second \underline{underlined}.}
	\label{tab:libero_comparison}
	\setlength{\tabcolsep}{12pt}
	\begin{tabular}{lccccccc}
		\toprule
		\textbf{Method} & \textbf{Scale} & \textbf{Params (B)} & \textbf{Spatial} & \textbf{Object} & \textbf{Goal} & \textbf{Long} & \textbf{Average} \\
		\midrule
		FlowVLA~\cite{flowvla} & \multirow{9}{*}{Large} & 8.5 & 93.2 & 95.0 & 91.6 & 72.6 & 88.1 \\
		UnifiedVLA~\cite{unifiedvla} & & 8.5 & 95.4 & \underline{98.8} & 93.6 & 94.0 & 95.5 \\
		OpenVLA~\cite{openvla} & & 7 & 84.7 & 88.4 & 79.2 & 53.7 & 76.5 \\
		OpenVLA-OFT~\cite{openvlaoft} & & 7 & 97.6 & {98.4} & {97.9} & {94.5} & \underline{97.1} \\
		UniVLA~\cite{univla} & & 7 & 96.5 & 96.8 & 95.6 & 92.0 & 95.2 \\
		CoT-VLA~\cite{cotvla} & & 7 & 87.5 & 91.6 & 87.6 & 69.0 & 81.1 \\
		WorldVLA~\cite{worldvla} & & 7 & 87.6 & 96.2 & 83.4 & 60.0 & 81.8 \\
		ThinkAct~\cite{thinkact} & & 7 & 88.3 & 91.4 & 87.1 & 70.9 & 84.4 \\
		MemoryVLA~\cite{memoryvla} & & 7 & {98.4} & 98.4 & 96.4 & \underline{95.6} & 96.5 \\
		\midrule
		4D-VLA~\cite{4d-vla} & \multirow{7}{*}{Medium} & 4 & 88.9 & 95.2 & 90.9 & 79.1 & 88.6 \\
		SpatialVLA~\cite{spatialvla} & & 4 & 88.2 & 89.9 & 78.6 & 55.5 & 78.1 \\
	$\pi_0$~\cite{pi_zero} & & 3 & 96.8 & \underline{98.8} & 95.8 & 85.2 & 94.2 \\
	$\pi_0$-FAST~\cite{pifast} & & 3 & 96.4 & 96.8 & 88.6 & 60.2 & 85.5 \\
	$\pi_{0.5}$~\cite{pi_zero_five} & & 3 & \underline{98.8} & 98.2 & \underline{98.0} & 92.4 & 96.9 \\
	SmolVLA~\cite{smolvla} & & 2.2 & 93.0 & 94.0 & 91.0 & 77.0 & 88.8 \\
		GR00T N1~\cite{gr00t} & & 2 & 94.4 & 97.6 & 93.0 & 90.6 & 93.9 \\
		\midrule
		VLA-OS~\cite{vlaos} & Efficient & 0.5 & 87.0 & 96.5 & 92.7 & 66.0 & 85.6 \\
		\midrule
		\textbf{StreamVLA (ours)} & Medium & 3 & \textbf{99.2} & \textbf{99.4} & \textbf{98.6} & \textbf{96.6} & \textbf{98.5} \\
		\bottomrule
	\end{tabular}
\end{table*}


\subsubsection{RoboTwin 2.0 Benchmark Comparison}

We compare StreamVLA against five strong baselines on RoboTwin 2.0 (ACT, DP3, RDT, $\pi_0$, $\pi_{0.5}$) in Table~\ref{tab:robotwin_comparison}.

\begin{table*}[t]
    \centering
    \small
    \caption{\textbf{Comparison on RoboTwin 2.0 benchmark.} We report success rates (\%) on representative dual-arm manipulation tasks with 100 trials per task. Each task has Easy and Hard variants. Easy uses fixed environment configurations without randomization; Hard applies aggressive domain randomization. Best results in \textbf{bold}, second best \underline{underlined}.}
    \label{tab:robotwin_comparison}
    \begin{tabular}{lcccccccccccc}
        \toprule
        \multirow{2}{*}{\textbf{Task}} & \multicolumn{2}{c}{\textbf{RDT~\cite{rdt}}} & \multicolumn{2}{c}{\textbf{ACT~\cite{act}}} & \multicolumn{2}{c}{\textbf{DP3~\cite{dp3}}} & \multicolumn{2}{c}{\textbf{$\pi_0$~\cite{pi_zero}}} & \multicolumn{2}{c}{\textbf{$\pi_{0.5}$~\cite{pi_zero_five}}} & \multicolumn{2}{c}{\textbf{StreamVLA (Ours)}} \\
        \cmidrule(lr){2-3} \cmidrule(lr){4-5} \cmidrule(lr){6-7} \cmidrule(lr){8-9} \cmidrule(lr){10-11} \cmidrule(lr){12-13}
        & Easy & Hard & Easy & Hard & Easy & Hard & Easy & Hard & Easy & Hard & Easy & Hard \\
        \midrule
        Beat Block Hammer & \underline{77} & \underline{37} & 56 & 3 & 72 & 8 & 43 & 21 & \underline{74} & \underline{38} & \textbf{79} & \textbf{42} \\
        Blocks Ranking RGB & 3 & 0 & 1 & 0 & 3 & 0 & \underline{19} & \underline{5} & \textbf{22} & \underline{10} & \textbf{22} & \textbf{12} \\
        Blocks Ranking Size & 0 & 0 & 0 & 0 & 2 & 0 & \underline{7} & \underline{1} & \textbf{18} & \underline{8} & \textbf{18} & \textbf{9} \\
        Click Alarmclock & 61 & 12 & 32 & 4 & \underline{77} & \underline{14} & 63 & 11 & \underline{79} & \underline{35} & \textbf{82} & \textbf{38} \\
        Dump Bin Bigbin & 64 & 32 & 68 & 1 & \underline{85} & \underline{53} & 83 & 24 & \underline{86} & \underline{60} & \textbf{89} & \textbf{64} \\
        Handover Mic & 90 & \underline{31} & 85 & 0 & \textbf{100} & 3 & \underline{98} & 13 & 97 & \underline{52} & \underline{98} & \textbf{58} \\
        \midrule
        \textbf{Average} & 55.0 & \underline{26.0} & 50.8 & 7.4 & \underline{65.8} & 10.0 & 55.6 & 25.1 & 62.7 & 33.8 & \textbf{71.3} & \textbf{37.2} \\
        \bottomrule
    \end{tabular}
\end{table*}


\noindent\textbf{Robustness to Domain Shifts.} 
On the \texttt{Easy} split (standard environment), StreamVLA achieves a {71.3\%} success rate, outperforming the strong RDT baseline (55.0\%) by a large margin. 
However, the most critical insight comes from the \texttt{Hard} split, which applies aggressive domain randomization (lighting, textures, camera poses). 
While standard policies suffer catastrophic drops (e.g., ACT drops $50.8\% \to 7.4\%$), StreamVLA maintains robust performance at {37.2\%}, surpassing the second-best method (RDT) by {+11.2\%}. 
This suggests that our \textit{foresight-driven gating} helps the policy generalize beyond superficial visual statistics by anchoring control to stable semantic goals.

\noindent\textbf{Emergent Spatial Reasoning.} 
Tasks like \textit{Blocks Ranking} (sorting by RGB or Size) require understanding abstract concepts that go beyond simple pick-and-place. 
Traditional baselines (ACT) fail completely ($0-1\%$) as they lack strong language grounding, while $\pi_0$ achieves only $7\%$ (Easy) and $\pi_{0.5}$ achieves $20\%$ (Easy average). 
In contrast, StreamVLA achieves {20\%} (Easy average). 
Although these tasks remain challenging, our model demonstrates comparable strong reasoning capabilities alongside $\pi_{0.5}$, leveraging its System 2 to decompose the abstract instruction ``Rank by size'' into concrete visual sub-goals grounded by completion-state imagination.

\subsubsection{Real-World Evaluation}

\begin{table}[t]
    \centering
    \caption{\textbf{Real-World Success Rates.} We conduct 20 trials for each task. \textbf{Spelling}: Long-horizon logic; \textbf{Insertion}: High-precision geometry; \textbf{Interf. Spelling}: Spelling task with active human interference.}
    \label{tab:realworld_results}
    \small
    \setlength{\tabcolsep}{6pt}
    \begin{tabular}{lccc}
        \toprule
        \textbf{Method} & \textbf{Spelling} & \textbf{Insertion} & \textbf{Interf. Spelling} \\
        \midrule
        OpenVLA-OFT~\cite{openvlaoft} & 40\% & 35\% & 10\% \\
        $\pi_{0.5}$~\cite{pi_zero_five} & 45\% & 35\% & 15\% \\
        WALL-OSS~\cite{walloss} & 15\% & 25\% & 5\% \\
        \textbf{StreamVLA (Ours)} & \textbf{90\%} & \textbf{70\%} & \textbf{55\%} \\
        \bottomrule
    \end{tabular}
\end{table}


\noindent\textbf{1. Long-Horizon Sequential Reasoning (``Spelling Bee'').} 
StreamVLA achieves a 90\% success rate by utilizing System 2 to maintain a global task plan. While the reactive baseline $\pi_{0.5}$ can handle individual pick-and-place actions (45\%), it lacks the temporal logic to sequence multiple letters correctly, often leading to repetitive errors.

\noindent\textbf{2. Fine-Grained Geometric Control (Letter Insertion).} 
Precision is the primary bottleneck here. StreamVLA's imagined completion image serves as a {spatial anchor}, guiding the 50Hz System 1 controller to achieve 70\% success. Without this high-level guidance, $\pi_{0.5}$ and OpenVLA frequently fail to align the blocks with the narrow slots, dropping to 35\%.

\noindent\textbf{3. Robustness to Dynamic Interference (Interf. Spelling).} 
This task evaluates the system's resilience to active human interference, where previously placed letters are forcibly moved or removed during execution. 
Such interference inherently violates the current sub-task's logical progression and alters the scene's semantic state. 
While standard reactive policies (success rate $\leq$ 15\%) lack the temporal awareness to re-sequence their actions, StreamVLA maintains a robust {55\% success rate}. 
This is facilitated by our {Gated Reasoning Mechanism}: the physical displacement of objects creates a significant discrepancy between the current observation and the previously imagined completion state. 
This visual-semantic mismatch automatically triggers the ``unlocking'' of System 2, which re-detects the sub-task boundary and generates a rectified plan to recover from the interference, demonstrating an emergent capability for closed-loop error correction.


\subsection{Ablation Study}
\label{subsec:ablation}

\begin{table*}[t]
    \centering
    \caption{Component Analysis. We compare StreamVLA against variants with specific components removed or modified. Success Rate (\%) is reported. Avg. Latency refers to the average wall-clock time per control step. w/o Gating: Performs full reasoning at every step. Fixed-Step Pred: Predicts $t+\Delta t$ instead of the sub-task completion state.}
    \label{tab:ablation}
    \small
    \setlength{\tabcolsep}{8pt}
    \begin{tabular}{lccc}
        \toprule
        \textbf{Variant} & \textbf{LIBERO-Long} & \textbf{RoboTwin-Hard} & \textbf{Avg. Latency (ms)} \\
        \midrule
        StreamVLA (Full) & 96.6 & 37.2 & 128 \\
        \midrule
        \textit{A. Impact of Gating} & & & \\
        \quad w/o Gating (Always Reason) & 96.8 & 38.1 & 244 \\
        \quad w/o System 2 ($\pi_{0.5}$ only) & 92.4 & 33.8 & 65 \\
        \midrule
        \textit{B. Impact of Modalities} & & & \\
        \quad w/o Visual Imagination & 95.1 & 35.4 & 118 \\
        \quad w/o Textual Planning & 93.5 & 34.6 & 96 \\
        \midrule
        \textit{C. Impact of Prediction Target} & & & \\
        \quad Fixed-Step Pred ($t+\Delta t$) & 96.2 & 35.7 & 244 \\
        \bottomrule
    \end{tabular}
\end{table*}

\begin{figure}[t]
\centering
\includegraphics[width=\linewidth]{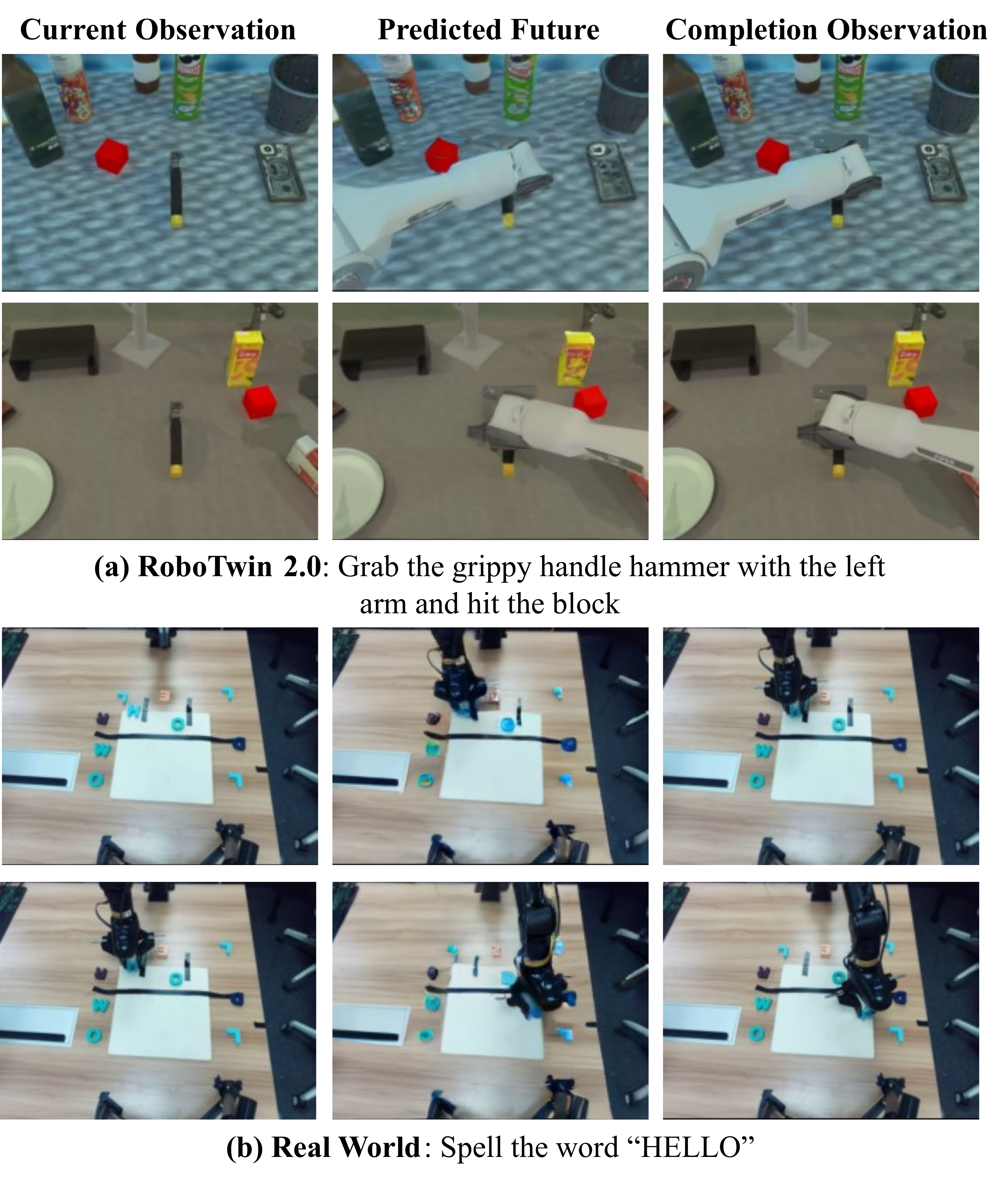}
\caption{\textbf{Future-completion prediction examples.} (a) RoboTwin 2.0 bimanual hammer task. (b) Two real-world Piper spell task. Each row shows current observation, predicted and ground future-completion frame.}
\label{fig:foresight_vis}
\end{figure}

\noindent\textbf{1. Efficacy of the Gating Mechanism.} 
As shown in Table~\ref{tab:ablation} (Row A), our gating mechanism achieves a Pareto-optimal balance between performance and speed. While the {w/o Gating (Always Reason)} variant provides the highest success rate, the improvement over our full model is marginal ($<1\%$). However, it incurs a severe latency penalty, increasing the average response time from 128ms to 244ms. This suggests that continuous re-planning reaches a point of diminishing returns, and our gating strategy successfully captures the critical ``decision points'' for System 2 without the overhead of redundant computation.

\noindent\textbf{2. Synergy of Text and Visual Imagination.} 
The ablation in Row B confirms that both textual and visual components are vital, but play distinct roles. Notably, removing either modality still results in higher performance than the $\pi_{0.5}$-only baseline, demonstrating that even partial System 2 reasoning provides valuable guidance.
\begin{itemize}
    \item \textbf{Textual Planning} appears more critical for long-horizon tasks. Removing it (``w/o Textual Planning'') leads to a larger drop on \texttt{LIBERO-Long} (96.6\% $\to$ 93.5\%), as the policy loses the explicit temporal logic required to sequence sub-tasks.
    \item \textbf{Visual Imagination} acts as a spatial anchor for precision. Its absence (``w/o Visual Imagination'') primarily impacts the perception-heavy \texttt{RoboTwin-Hard} benchmark, where the imagined goal image provides essential spatial grounding for end-effector control. 
\end{itemize}

\noindent\textbf{3. Completion State vs. Fixed-Step Prediction.} 
Row C highlights the importance of the prediction target. Modifying the head to predict a fixed future offset ($t+\Delta t$) instead of the completion state results in a performance decline across both benchmarks. Despite having the same high latency as the ``Always Reason'' variant (244ms), it fails to match its accuracy. This supports that \textit{what} the model predicts is as important as \textit{when} it predicts; our completion-state target provides a time-invariant goal that is more robust to variations in robot execution speed.

\subsection{Qualitative Analysis}
\label{subsec:qual}

\noindent\textbf{Visual Foresight and Execution Alignment.} 
Figure~\ref{fig:foresight_vis} illustrates the generation quality of our imagination head and its alignment with actual robot execution. Instead of merely matching a pre-defined static template, our model generates future-completion images that serve as semantic targets for the policy.

\noindent\textbf{Adaptive gating patterns.}
We observe a clear bimodal distribution: skip ratios are low (5-15\%) during critical {sub-task transitions}, ensuring frequent re-planning when uncertainty is high. 
Conversely, during {transport phases}, the skip ratio rises to 70-90\%, demonstrating that the model correctly identifies ``steady states'' where heavy reasoning is redundant.




\section{Conclusion}
\label{sec:conclusion}


In this work, we introduced StreamVLA, a unified framework that harmonizes the conflict between \textit{slow, deliberate reasoning} and \textit{fast, reactive control} in robotic manipulation. 
By embedding a dual-system architecture within a shared VLA backbone, we demonstrated that high-level planning does not need to come at the cost of inference latency. 
Our key contributions—the {future-completion imagination} head and the \textbf{dynamic gating} mechanism—enable the policy to maintain long-horizon coherence while executing at {50Hz}, effectively amortizing the cost of reasoning to achieve a {48\% latency reduction}. 
Extensive evaluations on LIBERO, RoboTwin, and real-world platforms validate that StreamVLA achieves state-of-the-art performance, offering a scalable path for deploying large multimodal models in dynamic, real-time physical systems. 

\noindent\textbf{Limitations and Future Work.}
Despite its robustness, our approach has limitations. 
Despite its robustness, our approach is limited by potential visual hallucinations under heavy occlusion and the reliance on a heuristic gating threshold $\tau$.
Looking ahead, we plan to extend the ``imagination'' paradigm beyond vision to include {tactile and force expectations}. 
Since our architecture is modality-agnostic, incorporating haptic feedback would enable the system to reason about contact-rich manipulation tasks where visual feedback alone is insufficient.

\bibliographystyle{plainnat}
\bibliography{references}

\clearpage
\appendix

\section*{Supplementary Material}

This supplementary material provides additional details on the implementation, training process, and experimental setups of StreamVLA.

\section{Implementation Details}
\label{app:implementation}

\subsection{Network Architecture}
\noindent\textbf{Shared Backbone.} 
The core of StreamVLA is its \textbf{shared backbone}, a unified architecture designed to bridge the gap between high-level reasoning and low-level control. Building upon the $\pi_{0.5}$~\cite{pi_zero_five} framework, the model integrates a \textbf{Vision Encoder} to transform multi-view observations into visual tokens. These tokens are subsequently processed alongside natural language instructions and proprioceptive states by a language model, which serves as the primary processing unit for multimodal input. This shared backbone ensures that the vast majority of parameters are utilized across both System 2's autoregressive reasoning and System 1's continuous action generation, significantly enhancing parameter efficiency.

\begin{itemize}
    \item \textbf{Vision Encoder:} The vision encoder is based on the SigLIP model, which has been pretrained with large-scale contrastive learning. It processes images at a resolution of $224 \times 224$, encoding them into a sequence of visual tokens. These visual tokens are concatenated with text tokens and fed into the unified transformer backbone.
    
    \item \textbf{Language Model:} The language model used in StreamVLA is a \textbf{3B parameter transformer}. The model processes both the visual tokens (from the vision encoder) and text tokens together, applying full bidirectional attention over both input modalities. Causal masking is applied only to the subtask prediction tokens to ensure that tasks are predicted sequentially.
\end{itemize}

\noindent\textbf{Subtask Head (System 2).}
The Subtask Head in StreamVLA reuses the language model from the shared backbone. During training, the loss is computed solely for the subtask portion, ensuring efficient optimization. In inference, the Subtask Head autoregressively generates the textual subtask instructions. This design enables effective sub-task instruction generation while minimizing computational overhead, as it leverages the pre-trained backbone and avoids redundant autoregressive decoding during steady-state execution.

\noindent\textbf{Imagination Head (System 2).}
Following VLA-OS~\cite{vlaos}, we adopt the Infinity~\cite{infinity} architecture as a dedicated autoregressive generation head for future-completion imagination. The head generates a high-fidelity completion image in the coarse-to-fine decoding paradigm of VAR~\cite{var}, using a multi-scale schedule that progressively increases the spatial resolution (from $1{\times}1$ up to $14{\times}14$). We tokenize images with a discrete VQ-GAN tokenizer with a codebook size of 8192, and implement the generator with a 1024-d embedding and an 18-block, 8-head transformer decoder. Overall, the imagination head introduces approximately 312M additional parameters on top of the shared backbone.

\noindent\textbf{Semantic Gating Module.} 
The Gating Module acts as a lightweight temporal conductor that evaluates the alignment between the current state and the sub-goal using a Cross-Attention mechanism. It takes the head-mounted observation $o_t^{head}$ as the Query and the locked completion state $\hat{o}_{locked}^{future}$ as the Key/Value pair, utilizing linear matrices $W_Q, W_K, W_V \in \mathbb{R}^{d_{model} \times d_{attn}}$ to project features from the 3B shared backbone. The resulting output is fused with the subtask instruction embedding $e_t$ to produce a visual-semantic feature $h_t$, which is subsequently processed via Global Average Pooling and a 3-layer MLP with a hidden dimension of 256 to compute the Discrepancy Score: $d_t = \sigma(\text{MLP}(\text{GlobalPool}(h_t)))$. Based on a threshold $\tau=0.5$, this score determines the transition between System 1 and System 2.

\noindent\textbf{Flow Matching Action Expert (System 1)}
We employ Conditional Flow Matching (CFM) for action generation. The CFM model enables the precise prediction of motor control commands, facilitating efficient and accurate robotic movement.

The action generation process relies on the following key components:
\begin{itemize}
    \item \textbf{ODE Solver:} The action generation is driven by the Euler solver with a fixed step size to ensure stable and predictable behavior during trajectory computation.
    \item \textbf{Inference Steps:} During inference, we execute \textbf{10 steps} of ODE integration to generate the action chunk. This stepwise integration allows for smooth trajectory planning and efficient execution.
    \item \textbf{Action Chunk Size:} The action chunk size is set to $K=10$ for LIBERO benchmarks, providing precise control over action duration. For RoboTwin benchmarks and real-world tasks, we use a larger chunk size of $K=50$ to accommodate the more complex joint movements in dynamic environments.
    \item \textbf{Action Space:} The action space is defined differently depending on the task. For LIBERO, the action space consists of the relative end-effector pose (6D) and the gripper state (1D). For both real-world and RoboTwin tasks, the action space includes the absolute joint positions (6D) and the gripper state (1D). In the case of RoboTwin, which involves dual-arm control, the action space includes joint positions for both arms (12D) and the gripper state for each arm (2D).
\end{itemize}

\noindent\textbf{Parameter Breakdown.} Table~\ref{tab:params} shows the complete parameter distribution.

\begin{table}[h]
\centering
\small
\caption{Parameter breakdown across model components.}
\label{tab:params}
\begin{tabular}{lcc}
\toprule
\textbf{Component} & \textbf{Params} & \textbf{\%} \\
\midrule
Vision Encoder & 412M & 12.47\% \\
Language Backbone & 2,508M & 75.89\% \\
Imagination Head & 312M & 9.44\% \\
Action Expert & 15M & 0.45\% \\
Gating Module & 58M & 1.75\% \\
\midrule
\textbf{Total} & \textbf{3,305M} & \textbf{100\%} \\
\bottomrule
\end{tabular}
\end{table}


\subsection{Training Configuration}\label{app:training}
We train StreamVLA using a two-stage curriculum strategy on a cluster of 24$\times$NVIDIA A800 GPUs with BF16 precision. In the first stage (Imagination Alignment), we freeze the pretrained VLM backbone and the Action Head, training only the Imagination and Subtask Heads to extract valid goal representations from the frozen latent space. In the second stage (Full Fine-tuning), we unfreeze all modules—including the backbone, Action Head, and autoregressive heads—for end-to-end joint optimization. Table~\ref{tab:hyperparams} lists the detailed hyperparameters and loss weights.

\begin{table}[h]
\centering
\small
\caption{Complete training hyperparameters for StreamVLA.}
\label{tab:hyperparams}
\begin{tabular}{lc}
\toprule
\textbf{Hyperparameter} & \textbf{Value} \\
\midrule
\multicolumn{2}{l}{\textit{Optimization}} \\
Stage I learning rate & \texttt{5e-5} \\
Stage II learning rate & \texttt{1e-5} \\
Optimizer & AdamW \\
Precision & BF16 / DDP \\
Gradient clipping (max norm) & \texttt{1.0} \\
\midrule
\multicolumn{2}{l}{\textit{Training Schedule}} \\
Stage I steps & 20K \\
Stage II steps & 40K \\
LR schedule & Cosine decay \\
Total batch size & 128 \\
\midrule
\multicolumn{2}{l}{\textit{Loss Weights}} \\
$\lambda_{sub}$ (Subtask Head) & \texttt{0.1} \\
$\lambda_{img}$ (Imagination Head) & \texttt{0.1} \\
$\lambda_{gate}$ (Gating Module) & \texttt{1.0} \\
$\mathcal{L}_{act}$ (Action Generation) & \texttt{1.0} \\
\midrule
\multicolumn{2}{l}{\textit{Image Data Augmentation}} \\
Random crop & 95\% \\
Random rotation & $\pm 5^\circ$ \\
Color jitter (Brightness/Contrast) & 0.7$\sim$1.3$\times$ \\
\bottomrule
\end{tabular}
\end{table}

\noindent\textbf{Stage I: Imagination Alignment.} We freeze the VLM backbone and the Action Head to force the autoregressive (AR) heads to learn goal-directed reasoning without disrupting pretrained spatial knowledge. This stage focuses on aligning the subtask text and visual completion states.

\noindent\textbf{Stage II: Full Fine-tuning.} All modules are unfrozen to allow the backbone to adapt its representations for both high-level reasoning (System 2) and reactive control (System 1). The gating module is supervised using binary cross-entropy (BCE) loss against transition labels derived from expert demonstrations.


\subsection{Data Annotation Pipeline}
\label{sec:annotation}

To train the \textit{Subtask Head} and \textit{Imagination Head}, we require dense supervision signals that align high-level semantic instructions with precise temporal boundaries and visual outcomes. 
However, standard benchmarks like \textbf{LIBERO}~\cite{libero} only provide language instructions for the global task (e.g., ``Put the cream cheese in the basket'') without intermediate subtask labels.
To address this, we construct a rigorous \textbf{Semi-Automated Annotation Pipeline} to upgrade existing demonstrations into a \textit{Chain-of-Thought (CoT)} format.

\noindent\textbf{1. VLM-Assisted Pre-Annotation.} 
We first utilize a state-of-the-art VLM Qwen3-VL-plus~\cite{qwen3-vl} to temporally segment the raw demonstration videos. 
We prompt the VLM to decompose the long-horizon task into logical sub-stages (e.g., \textit{Pick}, \textit{Place}, \textit{Retract}) and estimate the start and end timestamps for each segment. 
This provides a coarse-grained initialization of the subtask sequence.

\begin{figure*}[htbp]
    \centering
    \includegraphics[width=0.9\linewidth]{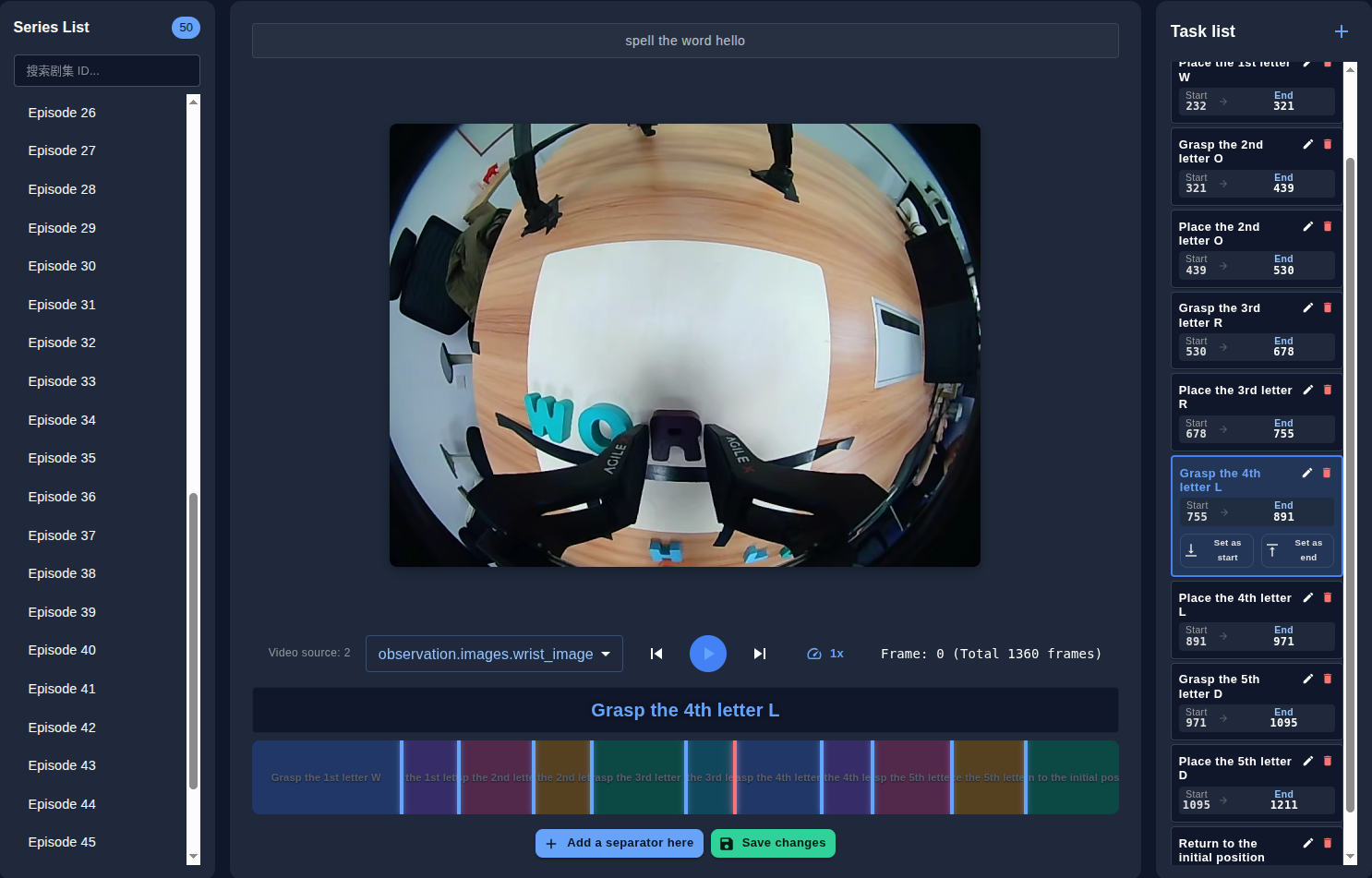} 
    \caption{\textbf{Semi-Automated Annotation Interface.} Our custom-built tool allows annotators to efficiently segment long-horizon videos, assign semantic instructions to subtasks, and verify the corresponding completion frames across diverse environments.}
    \label{fig:annotation_tool}
\end{figure*}

\noindent\textbf{2. Human Verification via Custom Tooling.} 
Since VLMs can suffer from temporal hallucinations or imprecise boundary detection, we developed a \textbf{custom web-based annotation tool} (see Figure~\ref{fig:annotation_tool}) for human verification. 
Expert annotators review the VLM-generated segments to:
(i) \textbf{Correct Logic:} Refine the text description $s_t$ to ensure it matches the physical action (e.g., changing "Move to mug" to "Grasp white mug").
(ii) \textbf{Refine Boundaries:} Precisely adjust the timestamp $t_{end}$ to the exact frame where the sub-goal is mechanically completed (e.g., the moment the gripper releases the object).
(iii) \textbf{Key-Frame Extraction:} The system automatically extracts the visual observation at $t_{end}$ as the ground-truth \textbf{Completion Image} ($\hat{o}^{future}_{GT}$) for training the Imagination Head.

\begin{figure*}[t]
\centering
\includegraphics[width=0.9\linewidth]{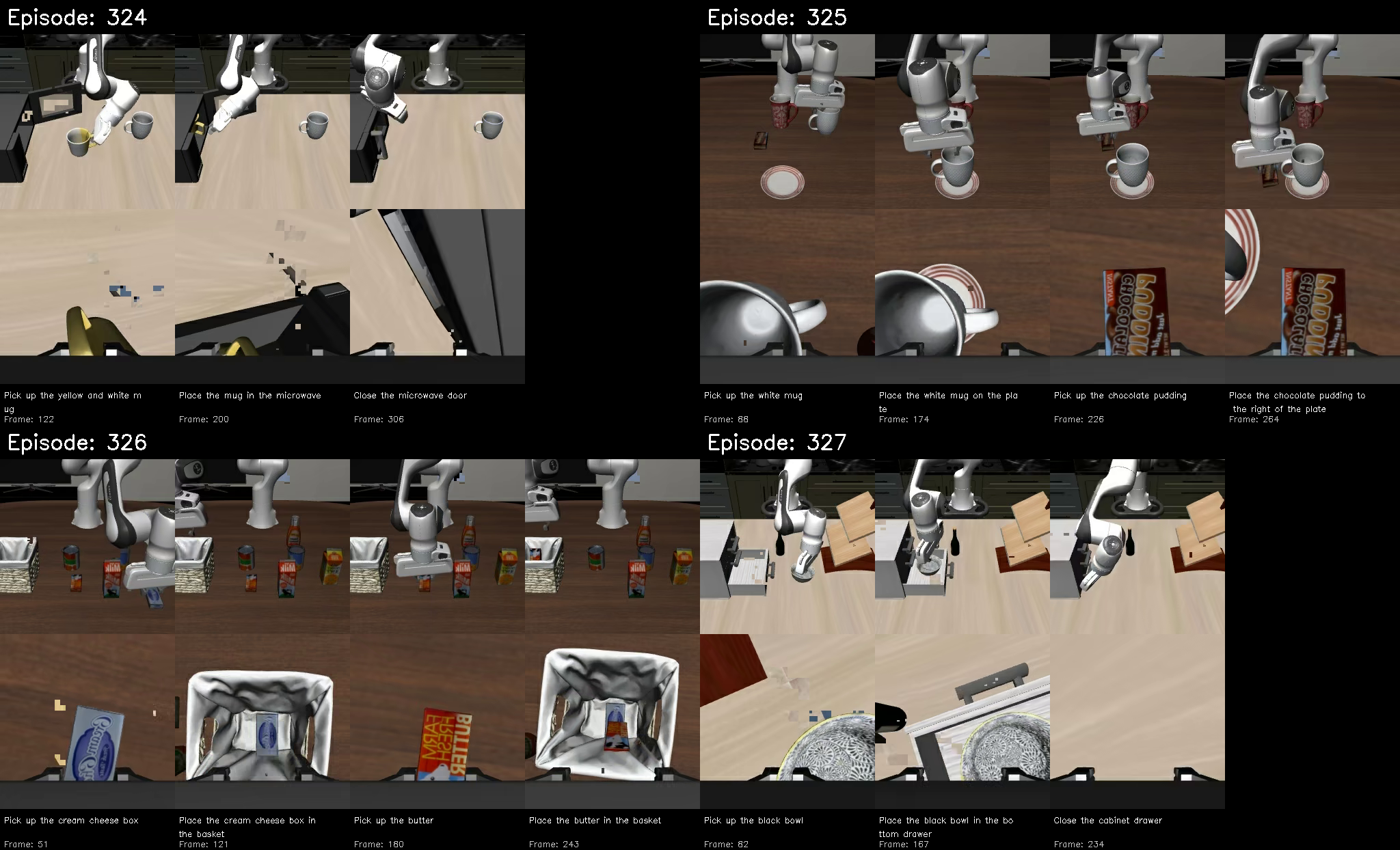}
\caption{\textbf{Annotated Subtask Sequences for Libero.}}
\label{fig:data_annotation_libero}
\end{figure*}

\begin{figure*}[t]
\centering
\includegraphics[width=0.9\linewidth]{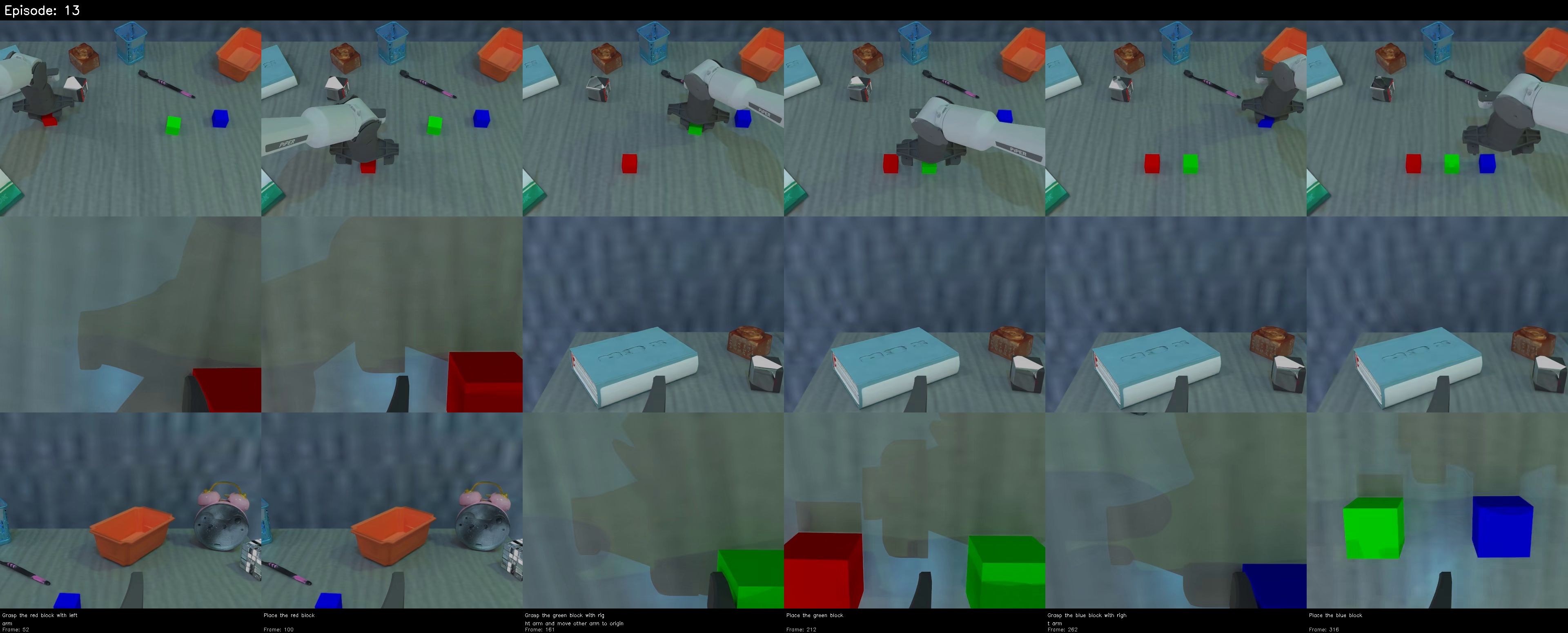}
\caption{\textbf{Annotated Subtask Sequences for RoboTwin.}}
\label{fig:data_annotation_robotwin}
\end{figure*}

\begin{figure*}[t]
\centering
\includegraphics[width=0.9\linewidth]{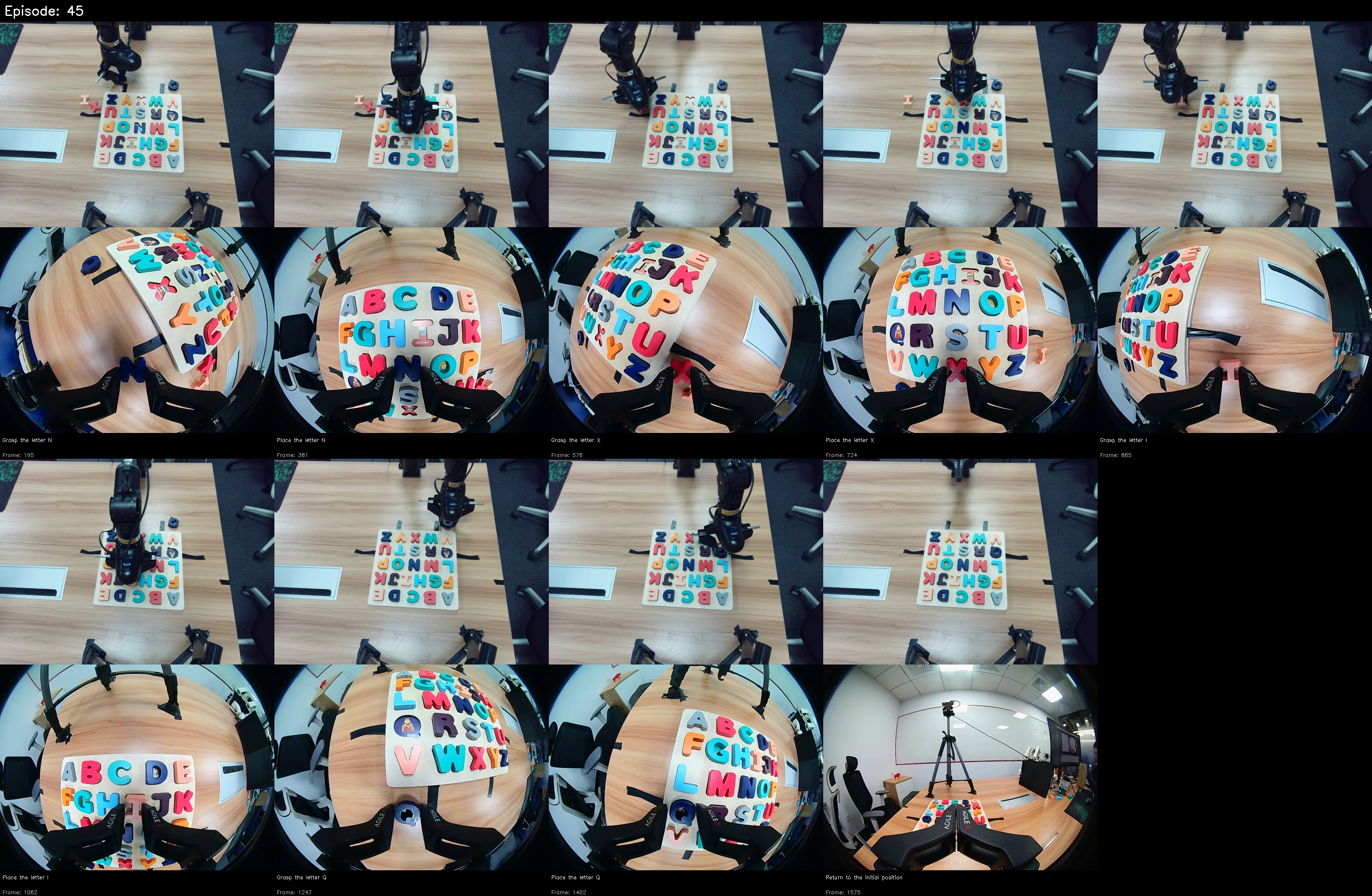}
\caption{\textbf{Annotated Subtask Sequences for Letter Insertion.}}
\label{fig:data_annotation_inserting}
\end{figure*}

\begin{figure*}[t]
\centering
\includegraphics[width=0.9\linewidth]{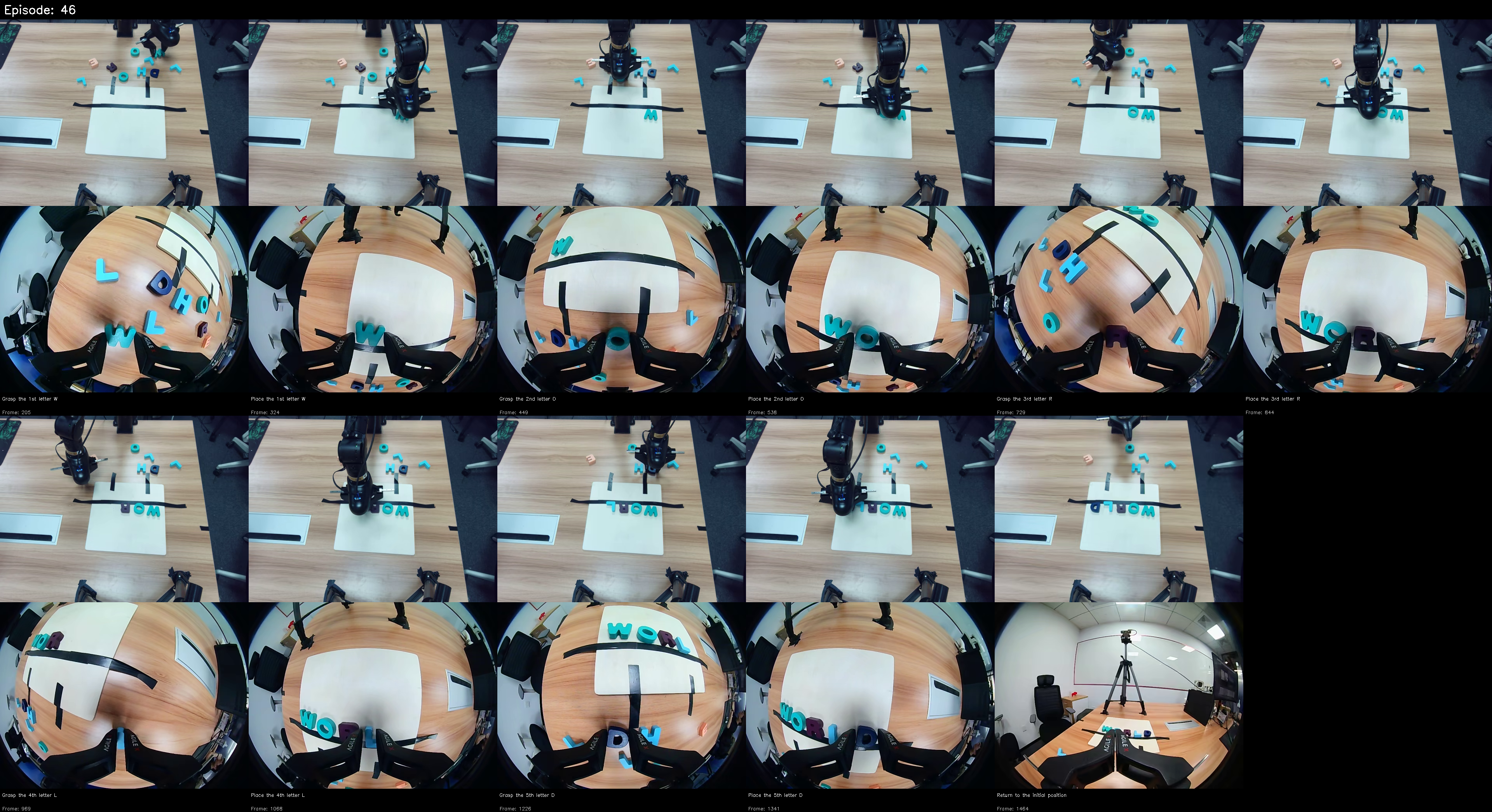}
\caption{\textbf{Annotated Subtask Sequences for Spelling Bee.}}
\label{fig:data_annotation_spelling}
\end{figure*}

\noindent\textbf{3. Outcome.} 
This semi-automated pipeline yields a high-quality, multimodal dataset where each sequential subtask is precisely paired with a semantic instruction and a corresponding \textit{completion frame}. As illustrated in Figures~\ref{fig:data_annotation_libero} to \ref{fig:data_annotation_spelling}, our annotation covers a diverse range of manipulation challenges across both simulation and physical environments:

\begin{itemize}
    \item \textbf{Simulation Benchmarks:} We decompose trajectories in \textbf{LIBERO} (Fig.~\ref{fig:data_annotation_libero}) and \textbf{RoboTwin} (Fig.~\ref{fig:data_annotation_robotwin}) into logical stages, such as ``Open microwave'' $\to$ ``Place mug'' $\to$ ``Close door'', providing a rich set of multi-stage interaction data.
    \item \textbf{Real-World Tasks:} For high-precision and logic-heavy tasks, we annotated sequences for \textbf{Letter Insertion} (Fig.~\ref{fig:data_annotation_inserting}), focusing on sub-centimeter alignment, and \textbf{Spelling Bee} (Fig.~\ref{fig:data_annotation_spelling}), which requires multi-step symbolic reasoning (e.g., ``Pick `S''' $\to$ ``Place `S''' $\to$ ``Pick `P''').
\end{itemize}

This dense temporal supervision provides the ground-truth completion states ($\hat{o}^{future}_{GT}$) and transition boundaries necessary to train the \textbf{Lock-and-Gated mechanism}. This ensures that the model can reliably identify sub-goal completion and effectively switch between System 2's high-level reasoning and System 1's reactive control.

\subsection{Real-World Experimental Details}
\label{app:real_world_setup}

\noindent\textbf{Hardware Configuration.}
\begin{itemize}
    \item \textbf{Robot Arm:} AgileX Piper (6-DoF lightweight arm).
    \item \textbf{Mobile Base:} Custom differential-drive chassis (for Stacking task).
    \item \textbf{Cameras:} 
        \begin{itemize}
            \item \textit{Head View:} Custom Webcam, mounted externally.
            \item \textit{Wrist View:} Custom fisheye RGB camera.
        \end{itemize}
    \item \textbf{Compute:} All inference runs on a generic workstation with a single NVIDIA RTX 4090 GPU to simulate edge deployment.
\end{itemize}

\noindent\textbf{Evaluation Protocol.}

\noindent\textbf{Task 1: Spelling Bee} 
\noindent\textit{Objective:} Sequence multiple actions to correctly spell the target word on the workspace.
\noindent\textit{Subtasks:}
\begin{itemize}
    \item Pick up the first letter block.
    \item Place the first letter block in the target position.
    \item Sequentially pick up and place the subsequent letter blocks according to the spelling order.
    \item Return the robotic arm to its initial home position after the word is completed.
\end{itemize}
\noindent\textit{Success criteria:} The word is correctly spelled, and the arm returns to its starting configuration.

\noindent\textbf{Task 2: Letter Insertion} 
\noindent\textit{Objective:} Achieve sub-centimeter accuracy by inserting specified letter blocks into narrow slots using the 6-DoF AgileX Piper robotic arm.
\noindent\textit{Subtasks:}
\begin{itemize}
    \item Pick up the target letter block.
    \item Insert the letter block into the designated narrow slot.
    \item Sequentially repeat the pick-and-insert process for all required blocks.
    \item Return the robotic arm to its initial home position upon completion.
\end{itemize}
\noindent\textit{Success criteria:} All target blocks are fully seated within their respective slots.

\vspace{0.5em}
\noindent\textbf{Task 3: Spelling Bee with Interference} 
\noindent\textit{Objective:} Successfully complete the spelling the target word despite active human interference. The subtasks are identical to Task 1, assessing the system's ability to handle semantic disruptions and re-plan in real-time.
\noindent\textit{Interference Scenarios:} A human operator is instructed to intervene 1--2 times per episode. Interventions include:
\begin{itemize}
    \item Moving the target object while the robot is reaching for it.
    \item Removing the target slot board or altering the workspace layout.
    \item Forcibly moving previously placed letter blocks to incorrect positions.
    \item Removing previously placed letter blocks from the workspace entirely.
\end{itemize}
\noindent\textit{Success criteria:} The word is correctly spelled, and the arm returns to its starting configuration.

\subsection{Additional Experiments}

\noindent\textbf{Gating Threshold Analysis}
We analyze the impact of the gating threshold $\tau$ on the LIBERO-Long benchmark. Table~\ref{tab:threshold} illustrates the trade-off between task success, computational efficiency, and reasoning frequency.

\begin{table}[h]
\centering
\small
\caption{Gating threshold analysis on LIBERO-Long. Skip ratio indicates the percentage of timesteps where System 2 reasoning is bypassed.}
\label{tab:threshold}
\begin{tabular}{lcccc}
\toprule
\textbf{Threshold $\tau$} & \textbf{SR (\%)} & \textbf{Latency (ms)} & \textbf{Skip Ratio (\%)} \\
\midrule
0.3 & 93.5 & 92  & 88 \\
0.4 & 95.1 & 110 & 81 \\
0.5 (Default) & 96.6 & 128 & 72 \\
0.6 & 96.7 & 165 & 54 \\
0.7 & 96.7 & 195 & 35 \\
\midrule
Always Skip & 92.4 & 65 & 100 \\
Never Skip & 96.8 & 244 & 0 \\
\bottomrule
\end{tabular}
\end{table}

\noindent\textbf{Analysis.} Setting $\tau = 0.5$ provides the optimal balance between performance and efficiency, achieving a 96.6\% success rate while skipping expensive autoregressive reasoning 72\% of the time. This results in a 48\% reduction in average latency (from 244ms to 128ms) compared to the "Never Skip" baseline. Lower thresholds ($\tau < 0.5$) skip reasoning too aggressively, which leads to performance degradation as the model misses critical subtask transitions. Conversely, higher thresholds ($\tau > 0.5$) trigger reasoning too frequently, significantly increasing inference latency with only marginal gains in success rate. The results demonstrate that our adaptive gating mechanism effectively identifies steady-state execution phases to amortize the cost of reasoning without sacrificing long-horizon coherence.

\noindent\textbf{Inference-Time Output Ablation}
We ablate different output modalities during inference to assess their necessity for task performance versus computational cost on the LIBERO-Long benchmark. The full model utilizes a hybrid head design, while variants selectively bypass specific autoregressive reasoning components within System 2.

\begin{table}[h]
\centering
\small
\caption{Inference-time output ablation on LIBERO-Long. }
\label{tab:outputs}
\begin{tabular}{lcc}
\toprule
\textbf{Configuration} & \textbf{SR (\%)} & \textbf{Latency (ms)} \\
\midrule
Full model (StreamVLA) & 96.6 & 128 \\
w/o Subtask Head & 92.0 & 97 \\
w/o Imagination Head  & 92.6 & 120 \\
w/o System 2  & 90.6 & 65 \\
\bottomrule
\end{tabular}
\end{table}

\noindent\textbf{Analysis.} 
The results indicate that the \textbf{Subtask Head} is the most critical component for long-horizon coherence; removing it reduces latency to 97ms but causes the success rate to plummet to 92.0\%. This significant degradation confirms that textual subtask descriptions provide the essential temporal logic required for sequencing complex manipulation stages. 

The \textbf{Imagination Head} also proves vital, as its removal drops performance to 92.6\% with 120ms latency. The absence of the imagined completion state primarily impacts tasks requiring high spatial precision, as the visual foresight serves as a stable geometric anchor for the action head. Notably, the performance drop in this inference-time setting is more pronounced than in training-time ablation, suggesting that the backbone develops a strong reliance on these semantic anchors during joint optimization.

Finally, the \textbf{w/o System 2} configuration (System 1 only) achieves the minimum latency of 65ms but suffers the most substantial degradation to 90.6\%. This baseline demonstrates that standard reactive policies, even when derived from a powerful 3B backbone, lack the necessary temporal awareness and goal-directed guidance provided by our adaptive dual-system architecture.

\noindent\textbf{Conclusion.} Our ablation confirms that both the Subtask Head and the Imagination Head are vital for long-horizon task success. While removing these components reduces the overhead of autoregressive decoding, the full StreamVLA model achieves the optimal Pareto-balance between reasoning depth and execution speed.

\end{document}